\pdfoutput=1

\documentclass[11pt]{article}

\usepackage[preprint]{acl}

\usepackage{times}
\usepackage{latexsym}
\usepackage{balance}
\usepackage[T1]{fontenc}

\usepackage[utf8]{inputenc}

\usepackage{microtype}

\usepackage{inconsolata}

\usepackage{graphicx}
\usepackage{adjustbox}
\usepackage{listings}
\usepackage[dvipsnames]{xcolor}
\usepackage{amsmath}
\usepackage{multirow, multicol}
\usepackage{array}
\usepackage{tabularray}

\definecolor{myyellow}{HTML}{eeba0b}

\newcommand{\opt}{OPT-6.7B}
\newcommand{\flan}{Flan-T5 Large}
\newcommand{\gpt}{GPT-2 Medium}
\newcommand{\gptt}{GPT-3.5-turbo}
\newcommand{\gptf}{GPT-4-turbo}

\title{Training Zero-Shot Generalizable End-to-End Task-Oriented Dialog System Without Turn-level Dialog Annotations}

\author{Adib Mosharrof \\
  University of Kentucky \\
  Lexington, KY, USA \\
  \texttt{adib.mosharrof@uky.edu} \\\And
  A.B. Siddique \\
  University of Kentucky \\
  Lexington, KY, USA \\
  \texttt{siddique@cs.uky.edu} \\}

\newcommand{\oursys}{$\mathsf{NL}\mbox{-}\mathsf{TOD}$}

\newcolumntype{L}{!{\vrule width 1.1pt}c}
\newcolumntype{R}{c!{\vrule width 1.1pt}}

\begin{document}
\maketitle
\begin{abstract}


Task-oriented dialogue (TOD) systems enable users to achieve their goals through natural language interactions. 
Traditionally, these systems have relied on turn-level manually annotated metadata, such as dialogue states and policy annotations, which are expensive, time-consuming, and often inconsistent or error-prone. This dependence limits the potential to leverage vast amounts of readily available conversational data for training TOD systems.
Additionally, a critical challenge in TOD system design is \emph{determining when and how to access and integrate information from external sources}. 
Current approaches typically expect this information to be provided alongside the dialogue context, rather than learning to identify and retrieve it autonomously.
While pre-trained large language models (LLMs) have been used to develop TOD systems, their potential to train such systems \emph{without laborious annotations remains largely unexplored}. 
This work employs multi-task instruction fine-tuning to create more efficient and scalable TOD systems that can effectively leverage natural language conversational data without manual annotations, while autonomously managing external information retrieval.
Our extensive experimental evaluations, using three diverse TOD datasets and three LLMs of varying sizes, demonstrate that our approach can generalize to new, unseen domains. 
Notably, our approach outperforms both state-of-the-art models trained on annotated data and billion-scale parameter off-the-shelf ChatGPT models.

\end{abstract}

\section{Introduction}

\begin{figure}[!t]
    \centering
    \includegraphics[width=0.97\linewidth]{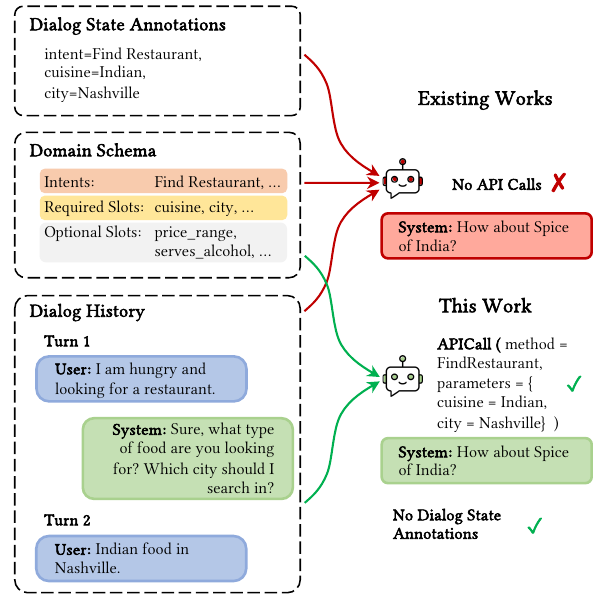}
    \vspace{-6pt}
    \caption{
This work leverages natural language conversational data alone, eliminating the need for extensive manual annotations required by existing works.
It achieves generalization to new, unseen domains through the use of domain schemas.}
    \label{fig:approach}
    \vspace{-14pt}
\end{figure}

Task-oriented dialogue (TOD) systems~\cite{zhang2020task} enable users to accomplish a wide range of goals through natural language interactions.
These systems power virtual assistants, customer service chatbots, and various other applications that facilitate tasks such as booking reservations, scheduling appointments, or retrieving information~\cite{Williams2016TheDS, Zhang2019TaskOrientedDS}. 
Traditionally, the development of TOD systems has relied heavily on turn-wise manually annotated metadata, including dialogue states and policy annotations~\cite{Zhang2020RecentAA}. This approach, while effective, presents significant challenges to scaling TOD systems to new domains, since this approach cannot exploit vast amounts of readily available raw conversational data for training.
That is, the process of creating these annotations is laborious and costly.
Moreover, the quality of these annotations can be inconsistent and prone to errors~\cite{eric2020multiwoz,zang2020multiwoz,han2021multiwoz,budzianowski2019challenges}.

Another critical challenge in the design of TOD systems is the effective integration of external information sources. 
Many real-world tasks require the system to access and incorporate information beyond the immediate dialogue context, such as querying databases and accessing knowledge bases. 
Current approaches often sidestep this challenge by assuming that the necessary external information is provided alongside the dialogue context. 
However, this assumption limits the autonomy and adaptability of these systems, as they cannot independently learn when external information is needed and how to retrieve it effectively from which source.

Recent advancements in natural language processing have seen the rise of pre-trained large language models (LLMs)~\cite{vaswani2017attention,devlin2019bert,radford2019language}, which have demonstrated remarkable capabilities across various language tasks. 
While these models have been applied to develop TOD systems~\cite{hosseini2020simple,Yang2020UBARTF,Mosharrof2023ZeroShotGE,budzianowski2018towards}, their potential to train such systems without the need for extensive manual annotations remains largely unexplored. 
This untapped potential presents an opportunity to enable the development of end-to-end (E2E) zero-shot generalizable TOD systems.

In this work, we propose to move away from manually annotated turn-wise data and present a novel \textbf{N}atural \textbf{L}anguage \textbf{T}ask \textbf{O}riented \textbf{D}ialog System, {\oursys}, that employs multi-task instruction fine-tuning~\cite{raffel2020exploring,wei2022finetuned,mishra2021cross} and uses only the dialog history and domain schema to create a Zero Shot Generalizable TOD system. 
Additionally, our approach integrates autonomous external information retrieval. 
Unlike existing systems that require external information to be provided explicitly, {\oursys} learns to identify when additional information is needed and how to retrieve it effectively. 
Our approach not only addresses the labor-intensive nature of traditional TOD system development but also opens up new possibilities for utilizing the wealth of unannotated dialogue data available in various domains.

To rigorously evaluate our approach, we conducted extensive experiments using three diverse TOD datasets: SGD~\cite{Rastogi2019TowardsSM}, KETOD~\cite{Chen2022KETODKT}, and BiToD~\cite{Lin2021BiToDAB}.
Our experiments employed three LLMs of varying sizes: {\gpt} (355M parameters)~\cite{radford2019language}, {\flan} (780M parameters)~\cite{chung2022scaling}, and {\opt} (6.7B parameters)~\cite{zhang2022opt}. 
For smaller models, we used full-model training, while for larger models, we applied the Low-Rank Adaptation (LORA)~\cite{Hu2021LoRALA} approach to investigate computational efficiency and zero-shot generalization capabilities. 
In addition to traditional result reporting metrics (e.g., BLEU, GLEU) for overall response generation, this work also examines performance at the sub-task level: request, inform, and API call capabilities (i.e., the accuracy of method names, parameters, parameter values, and exact full API calls), which provides a fine-grained assessment of the models' proficiency to fulfill users' intent while autonomously interacting with external systems.


Our results demonstrate that the proposed approach enables TOD systems to generalize effectively to new, unseen domains. 
Notably, our approach outperforms state-of-the-art (SOTA) models that were trained on manually annotated data. 
This finding challenges the conventional wisdom that high-quality annotations are necessary for achieving top performance in TOD tasks. 
Furthermore, our approach even surpasses the performance of billion-scale parameter off-the-shelf models (e.g., {\gptt}, {\gptf}~\cite{openai2023chatgpt}) in diverse TOD settings, despite using significantly smaller models.



\vspace{-5pt}
\section{Related Works}
\vspace{-5pt}

\textbf{Pipeline Approaches.}
TOD systems have traditionally been designed as pipeline systems, where separate components for Natural Language Understanding (NLU), Dialog State Tracking (DST), Dialog Policy, and Natural Language Generation (NLG) are used to handle specific parts of the dialog processing~\cite{ren2018towards, lee2013structured,peng2018deep,le2021predictable,wen2015semantically,peng2020few,chen2019semantically,budzianowski2018towards,Mosharrof2023TowardOS}.

\noindent
\textbf{End-to-End Approaches.}
Recent works have shifted towards E2E learning methods, where the TOD task is formulated as a conditional generation, where the model generates responses based on the entire dialog history and other relevant annotations (e.g., DST)~\cite{hosseini2020simple,Lin2021LeveragingSD,Bang2023TaskOptimizedAF,Zhang2023EnhancingPO,Ham2020EndtoEndNP,Chung2023InstructTODSLL,Yang2020UBARTF,Sun2022BORTBA,Imrattanatrai2023EndtoEndTD,Sun2022MarsSC,Zhao2022AnyTODAP,Peng2021SoloistBT,Mosharrof2023ZeroShotGE,Siddique2022PersonalizingTD}. 
For example, \citet{Lin2021LeveragingSD} introduced a slot description enhanced generative approach for zero-shot TOD, called T5DST. 
\citet{Zhang2023EnhancingPO} developed FiD-TOD that employs a caching mechanism for the dialog annotations, which can be extracted using a retrieval module.  

\noindent
\textbf{Prompting Approaches.}
Another recent research direction in TOD systems is in-context learning, where pre-trained LLMs are adapted to specific domains based on contextual examples without requiring fine-tuning~\cite{Labruna2023UnravelingCA,Hudevcek2023AreLL,Dingliwal2021FewSD, Madotto2020LanguageMA, Li2022ControllableDS, Madotto2021FewShotBP}.

\noindent
\textbf{Graph Based Approaches.}
Additionally, some works used graph-based methods to model the flow of dialog, e.g., by representing DST and policy decisions as graphs~\cite{Sohn2023TODFlowMT, Tuan2022TowardsLI, He2023KRPDSAK}. 
For instance, \citet{He2023KRPDSAK} introduced KRP-DS, a system that incorporates a knowledge graph as external information through a dedicated knowledge module, enabling context-aware path reasoning.
This graph assists in knowledge prediction, which is used to enhance response generation. 
\cite{Sohn2023TODFlowMT} introduced the TOD-Flow graph, where a graph is created from the dialog data and annotations, which uncovers the underlying task structure.

Most works in all categories require turn-level annotated data to train TOD models, which can be a significant limitation to scale in real world where such data may be scarce or costly.
Moreover, most systems assume that knowledge from external sources will be provided in the dialog context.
\emph{Our work, in contrast, focuses on training TOD models using only readily available natural language interactions, while leveraging domain schemas to enable scalability across unseen domains.}
\section{Methodology}
\vspace{-4pt}

\subsection{Problem Formulation}
\vspace{-2pt}
We formulate TOD response generation as a conditional sequence generation task. 
We use dialog history and domain schema to facilitate the creation of a zero-shot TOD model that can accommodate interactions from both seen and unseen domains.

Formally, in a multi-domain TOD setting, let \( D = \{d_1, d_2, \ldots, d_n\} \) be a set of domains, where each \( d_i \in D \) represents a distinct domain of interest.
For each domain \( d_i \), there exists a set of intents \( I_{d_i} \) such that \( I_{d_i} \subseteq \bigcup_{j=1}^n I_{d_j} \) is the set of possible intents associated with the domain \( d_i \). 
Each intent \( i_j \in I_{d_i} \) represents a specific user intent in domain \( d_i \).
For each intent \( i_j \in I_{d_i} \) there exists a set of slots \( S_{i_j} \) such that 
\( S_{i_j} \subseteq \bigcup_{k=1}^m S_{i_k} \)
is the set of possible slots associated with the intent \( i_j \), where each slot \( s_k \in S_{i_j} \) represents a slot relevant to the intent \( i_j \) in the domain \( d_i \).
So, a domain \( d_i \) can be represented as follows: 
\(
d_i = \left( I_{d_i}, \{ (i_j, S_{i_j}) \mid i_j \in I_{d_i} \} \right)
\).
For example, in the restaurant domain, the intent could be ``Book Table'' with relevant slots such as ``number\_of\_people'', ``date'', and ``restaurant\_name''.


Similarly, assume we have a dataset of
dialogs between a user and a system.
Each dialog \(\mathcal{T}_i\) consists of up to \( T \) turns, represented as a sequence:
\(
\mathcal{T}_i = \left( \{ (u_t, r_t) \mid t = 1, 2, \ldots, T \} \right)
\), where \( u_t \) and \( r_t \) represent the user's and system's utterance at turn \( t \).
Thus, the dialog history \(H_t\) at turn \(t\) consists of all the user and system utterances up user's utterance at turn \(t\), denoted as: 
\(
H_t = \{ (u_{1}, r_{1}),\dots, (u_{t-1}, r_{t-1}), u_{t} \}
\).
It is important to note that a dialog session may span up to \(m\) domains: $ \mathcal{T}_i \sim \bigcup_{j=1}^m d_{j} \subseteq  D$, where \(m \leq n \). 
At turn $t$, the system utterance $r_t$ can be generated by conditioning on the relevant domain schemas \(\bigcup_{j=1}^m d_{j}\) and dialog history $H_t$.


\subsection{Optimization of {\oursys}}

An autoregressive model (e.g., GPT-2~\cite{radford2019language}) generates text by predicting the next token given the preceding context.
For a given sequence of tokens \( (x_1, x_2, \ldots, x_{t-1}) \), the model predicts the probability distribution for the next token \( x_t \) as follows:
\(
p(x_t \mid x_{1:t-1}; \theta) = f_{\theta}(x_{1:t-1})
\), where \( f_{\theta} \) outputs a probability distribution over model's vocabulary \( \mathcal{V} \).
The next token \( x_t \) is sampled from the predicted probability distribution \( p(x_t \mid x_{1:t-1}; \theta) \).
This formulation readily extends to response generation for TOD:
\(
r_t \sim p(r_t \mid H_{t}; \theta)
\),
where the system response \(r_t\) for turn \(t\) is generated token-by-token by an iterative sampling.

In our formulation of {\oursys}, we introduce an additional conditioning variable domain schema $d_j \forall_{j}$ to encourage zero-shot generalization to unseen domains.
Then, the above formulation can be extended to optimize the model for the overall multi-turn dialog of \(T\) turns using a dataset of \(N\) dialogs with the following loss function:
\(
\mathcal{L}_{NL-TOD} = - \sum_{i=1}^{N} \sum_{t=1}^{T} \log p(r_{t} \mid H_{t}, d_{j}; \theta)
\).
In practice, dialog history \(H_t\) only consists of the last $k$ turns, where $k \leq t$, depending on the model's context length.

\subsection{Multi-task Instruction Fine-tuning}
\label{sec:response_breakdown}

The output of our system can be broadly categorized into two types: natural language responses to the user and API calls to external sources.
Our formulation trains the model for both outputs in an E2E fashion using multi-task instruction fine-tuning. 
It is important to note that we do not introduce any special tokens to tell the model which type of response to generate, instead the model learns autonomously when to generate API calls and when to generate a natural language response for the user.
In the following, we provide a brief overview of both types of responses.

\subsubsection{Response Generation}

The response generation is a natural language response for the user, which can be broken down into multiple sub-categories: request, inform, and general interaction.

\noindent
\textbf{Request.}
To fulfill user intents, the system may need additional information that the user has not yet provided.
For example, if a user asks to book a flight without specifying the date, the system would ask, ``What date would you like to travel?''.

\noindent
\textbf{Inform.}
The user may also need additional information to make informed decisions, which the system must provide. 
This involves retrieving and presenting requested details to the user.
For example, if a user asks, ``Does the restaurant have live music?'' the system must query the database and provide the requested information to the user.


\noindent
\textbf{General Interaction.}
The system also needs to engage in general interactions. 
This involves managing the flow of the conversation, offering help or suggestions, confirming user inputs, and providing general information that assists the user in navigating through the task. 

\subsubsection{API Calls}
\vspace{-3pt}
A critical aspect of TOD systems is their ability to communicate with external sources to retrieve additional information needed to complete a task. This category of system output involves making accurate API calls.
Consider a dialog scenario in the travel domain, where a user asks for Flights to New York the next Friday at 10 AM. 
Given the dialog context, the system must get a list of flights for recommendation, which requires generating an API call:
\texttt{APICall(method = FindFlight, parameters = \{destination=New York, date=next Friday, time=10AM, origin=CurrentLocation\})}.
Using the information returned from the API call, the system can suggest flights to the user.
It is important to note that the API call has several components: method name, list of parameters, and values associated with parameters.

In multi-domain settings, a special case arises when transferring information between domains, where the same data is carried forward but under different parameter names. 
For example, a parameter named \texttt{destination} in the travel domain might be referred to as \texttt{hotel\_city} in the subsequent hotel booking domain. 
An intelligent TOD model should autonomously perform this mapping.

\subsection{Training Details}
\vspace{-3pt}
The dialog history and domain schema are passed through a structured template to form the inputs to the model. 
This approach allows the model to generate both types of responses (i.e., natural language response, API call) efficiently by learning from instructions simultaneously. 
The template is detailed in Figure~\ref{fig:finetuning_template} in Appendix~\ref{sec:templates}.
%
{\oursys} leverages three pre-trained LLMs, namely {\gpt}, {\flan}, and {\opt}, and further refines them through multi-task instruction fine-tuning~\cite{raffel2020exploring,wei2022finetuned,mishra2021cross}. 
Several key optimization techniques were employed to achieve optimal training. 
An early stopping mechanism on the evaluation loss was implemented to prevent overfitting with the patience value set to 3.
The training process included 500 warm-up steps, during which the learning rate was gradually increased to stabilize the training process and avoid sudden jumps in the loss function. 
The AdamW~\cite{Loshchilov2017DecoupledWD} optimizer with a weight decay was used with a learning rate of 0.001. 
For {\gpt}, we employed full model training, while for {\flan} and {\opt}, we employed the Low-Rank Adaptation~(LoRA)~\cite{Hu2021LoRALA} and Quantized Low-Rank Adaptation~(QLoRA)~\cite{dettmers2023qlora} approach. 
LoRA facilitates parameter-efficient fine-tuning by injecting trainable low-rank matrices into each layer of the transformer architecture and quantization uses 8-bit integer~\cite{jacob2018quantization} as the data type to reduce the memory footprint.

\vspace{-4pt}
\section{Experimental Setup}
\vspace{-4pt}





\begin{table}
    \centering
    \small
    \begin{adjustbox}{max width=0.49\textwidth}
        \begin{tabular}{|l|c|c|c|c|c|c|}
            \hline
            Datasets $\rightarrow$ & \textbf{SGD} & \textbf{KETOD} & \textbf{BiToD}  \\ \hline
            \# Dialogs & 16,142  & 5,324 & 3,689 \\ \hline
            Average Turns / Dialog & 20.44   & 9.78 & 9.39 \\ \hline
            \# Domains & 20 & 14 & 5 \\ \hline
            \# Seen Domains & 16 & 12 & 5 \\ \hline
            \# Unseen Domains & 4 & 2 & 0 \\ \hline
            \# Unique API methods  & 46  & 38 &  8 \\ \hline
            \# Unique API parameters & 214 &  195 & 20 \\ \hline
        \end{tabular}
    \end{adjustbox}
    \vspace{-8pt}
    \caption{Important statistics of datasets.}
    \label{tab:data_statistics}
    \vspace{-14pt}
\end{table}

\subsection{Datasets}
\vspace{-3pt}
We use three diverse TOD datasets: Schema-Guided Dialog (SGD) dataset, Bilingual Task-Oriented Dialog (BiToD) dataset, and Knowledge-Enhanced Task-Oriented Dialog (KETOD) dataset. 
We have selected these datasets as these datasets define a schema to describe the domain and the conversational data contains API calls to communicate with external resources. Table~\ref{tab:data_statistics} shows detailed statistics about the datasets.

\vspace{-3pt}
\subsection{Evaluation Metrics}

To comprehensively evaluate the performance of {\oursys}, we separately assess individual sub-tasks and overall responses.
This approach can provide a more granular understanding of the system's strengths and areas for improvement.

\noindent
\textbf{Response Generation.}
We report BLEU-4~\cite{Papineni2002BleuAM} scores of overall system responses as well as request and inform sub-task as discussed in Section~\ref{sec:response_breakdown}. 
We also present the GLEU~\cite{wu2016googles} metric in Appendix~\ref{sec:appendix_gleu_mixed}.

\noindent
\textbf{API Calls.} We introduce a custom metric to assess different components of the API call. 
The API format is:
\(
\texttt{APICall}( \texttt{method} = method\_name,
             \texttt{parameters} = \{ (s_i,v_i)_{i=1}^n \} )
\)
The parameters attribute is a list of slot name and slot value pairs, where $s_i$ represents the slot name and $v_i$ represents the value of that slot. 
We use regular expressions to extract different parts of the API call to calculate the metrics.

            

\vspace{-6pt}
\begin{itemize}
\setlength\itemsep{-4pt}
    \item \textbf{Invoke Accuracy} measures whether the system can understand when to make an API call.
     \item \textbf{Method Accuracy} checks whether the appropriate method was used in the API call.      
     \item \textbf{Param Name Accuracy} assesses whether all the parameter names used to construct the API call are correct.      
     \item \textbf{Param Value Accuracy} evaluates whether each parameter value corresponding to a parameter name is correct. It is important to note that this metric will only be considered if the corresponding parameter name is correct. 
     \item \textbf{Complete API Call Accuracy} metric checks whether the complete API call (i.e., all components) was generated correctly.
\end{itemize}

            
\vspace{-15pt}
\subsection{Evaluation Settings}

To better understand the zero-shot generalization ability of {\oursys}, we evaluate the system on dialogs grouped by domain categories.
This evaluation is structured into four domain categories: All, Seen, Unseen, and Mixed. 

\noindent
\textbf{All Domains.} This category includes dialogs from all domains, providing a comprehensive evaluation of the system's overall performance.

\noindent
\textbf{Seen Domains.} This category contains dialogs from domains that were present in the training data. 
Evaluating seen domains demonstrates the supervised learning performance of the system.

\noindent
\textbf{Unseen Domains.} This category comprises dialogs from domains that were not included in the training data. 
Evaluating unseen domains tests the zero-shot generalizability of the system, demonstrating its ability to handle new, unseen domains without additional training.

\noindent
\textbf{Mixed Domains:} This category includes dialogs that span multiple domains, with some domains being seen and others unseen. 
This setting serves as an intermediary between supervised and zeo-shot settings. 
It tests the system's ability to manage dialogs where known and unknown domains are intermixed, reflecting more complex real-world scenarios. 
We present results for mixed domains in Appendix~\ref{sec:appendix_gleu_mixed}.

\begin{table*}[!t]
    \centering
    \begin{adjustbox}{max width=\textwidth}
        \begin{tabular}{|c|c|c|c| R c|c|R c|c|c|}
            \hline
    \multirow{2}{*}{\textbf{Dataset}}&\multirow{2}{*}{\textbf{Model}} & \multicolumn{3}{R}{\textbf{Overall Response (BLEU-4)}} & \multicolumn{3}{R}{\textbf{Inform (BLEU-4)}} & \multicolumn{3}{c|}{\textbf{Request (BLEU-4)}}  \\ \cline{3-11}
    & & \textbf{all}  & \textbf{seen} & \textbf{unseen} & \textbf{all}  & \textbf{seen} & \textbf{unseen} & \textbf{all}  & \textbf{seen} & \textbf{unseen} \\ \hline
    \multirow{4}{*}{SGD}&{\gptt} & 0.0397  & 0.0422 & 0.0363 
    & 0.1047 & 0.0998 & 0.1012
    & 0.0057 & 0.0046 & 0.0056 
    \\ 
    & {\gpt} & 0.1905  & 0.2150 & 0.1718 
    & 0.2227 & 0.2461 & 0.2059 
    & 0.0332 & 0.0350 & 0.0320
    \\ 
    & {\flan} & \underline{0.2680}  & \underline{0.2980} & \underline{0.2500} 
    & \underline{0.2991} & \underline{0.3064} & \underline{0.2872} 
    & \underline{0.0657} & \underline{0.0742} &0.0698
 \\ 
     & \opt & \textbf{0.3063}  & \textbf{0.3311} & \textbf{0.2896}
    & \textbf{0.3334} & \textbf{0.3499} & \textbf{0.3190}
    & \textbf{0.0836} & \textbf{0.0928} & \textbf{0.0793}
 \\  \hline
    
    \multirow{4}{*}{KETOD} & {\gptt} & 0.0397  & 0.0412 & 0.0334 
    & 0.0943 & 0.0801 & 0.0958
    & 0.0033 & 0.0036 & 0.0037
    \\ 
    & \gpt & 0.1509  & 0.1756 & 0.1224 
    & 0.1734 & 0.1843 & 0.1498
    & 0.0140 & 0.0193 & 0.0073
    \\ 
    & \flan &\underline{0.2130}  & \underline{0.2503} & \underline{0.1794}
    & \underline{0.2311} & \underline{0.2435} & \underline{0.2120}
    & \underline{0.0587} & \underline{0.0781} & \underline{0.0430}
 \\ 
    & \opt &\textbf{0.2680}  & \textbf{0.2994} & \textbf{0.2371} 
    & \textbf{0.2968} & \textbf{0.2981} & \textbf{0.3034} 
    & \textbf{0.0640} & \textbf{0.0851} & \textbf{0.0509} 
    \\ \hline

    \multirow{4}{*}{BiToD} & {\gptt} &0.0646 & 0.0646 & -
    & 0.2589 & 0.2589 & - 
    & 0.0011 & 0.0011 & - 
    \\ 
    & \gpt &\underline{0.3466}  & \underline{0.3466} & -
    & \underline{0.5226} & \underline{0.5226} & -
    & \underline{0.0323} & \underline{0.0323} & -
    \\ 
        & \flan &0.3159  & 0.3159 & -
    & 0.4795 & 0.4795 & -
    & 0.0256 & 0.0256 & -
     \\ 
 & OPT &\textbf{0.4143}  & \textbf{0.4143} & {-} 
    & \textbf{0.5248} & \textbf{0.5248} & {-} 
    & \textbf{0.0476} & \textbf{0.0476} & {-} 
 \\ \hline
        \end{tabular}
    \end{adjustbox}
    \vspace{-6pt}
    \caption{Results for natural language response generation and sub-tasks.}
    \label{tab:response_results}
    \vspace{-8pt}
\end{table*}

\begin{table}
    \centering
    \small
    \begin{adjustbox}{max width=0.45\textwidth}
        \begin{tabular}{|l|l|c|}
            \hline
            \textbf{Dataset}& \textbf{Model} & \textbf{BLEU-4} \\  \hline
            \multirow{6}{*}{SGD} & Soloist & 0.1120 \\ 
            &T5DST & 0.1700 \\ 
            &FiD-TOD & 0.1790 \\ 
            &ZS-TOD & 0.2000 \\ 
            & \opt & \textbf{\multirow{2}{*}{0.3063}} \\
            & \textbf{(this work)} & \\ \hline
            \multirow{4}{*}{KETOD}&Combiner & 0.099 \\ 
            & SimpleTODPlus & 0.1170 \\ 
            & \opt & \textbf{\multirow{2}{*}{0.2680}} \\
            & \textbf{(this work)}& \\ \hline
            \multirow{4}{*}{BiToD} & mbart & 0.3534 \\ 
            & mt5 & 0.3848 \\ 
            & \opt & \multirow{2}{*}{\textbf{0.4143}}  \\
            & \textbf{(this work)} &\\ \hline
        \end{tabular}
    \end{adjustbox}
    \vspace{-6pt}
    \caption{Comparison with SOTA approaches, which were trained using turn-level annotations. }
    \label{tab:sgd_sota}
    \vspace{-12pt}
\end{table}

\vspace{-4pt}
\subsection{Baselines}
\vspace{-4pt}
We establish several strong baselines for comparison to evaluate the effectiveness of \oursys.
Specifically, we compare \oursys with OpenAI's SOTA models and other SOTA approaches that use turn-level annotated dialog data. 

\noindent
\textbf{OpenAI's GPT Models.}
To compare with OpenAI's GPT models, we create a comprehensive prompt for the input dialog. 
The prompt is detailed in Figure~\ref{fig:chatgpt_template} in Appendix~\ref{sec:templates}. 
The template includes the dialog history, domain schema, and detailed instructions on how to process the dialog, understand the domain schema, and task-specific instructions.
We perform comprehensive comparisons with {\gptt} that uses this detailed prompt.
Moreover, we perform a limited-level comparison with the more expensive variant {\gptf}.

\noindent
\textbf{SOTA approaches.} We also compare {\oursys} against existing SOTA approaches. 
However, it is important to note that current SOTA methods do not report results for the API call task. Therefore, for these comparisons, we focus solely on the response generation task. 

\vspace{-6pt}
\begin{itemize}
\setlength\itemsep{-4pt}
    \vspace{-4pt}
    \item Soloist~\cite{Peng2021SoloistBT} introduced an E2E TOD system that employs a transformer-based autoregressive model that generates dialog responses grounded in user goals and real-world knowledge for task completion.

    \item T5DST~\cite{Lin2021LeveragingSD} introduced a slot description enhanced generative approach for zero-shot TOD.

    \item FiD-TOD~\cite{Zhang2023EnhancingPO} introduced a cache that contains accessible domains, intents, slots, and APIs. A retrieval module extracts relevant information from the cache and feeds the generative model for E2E TOD generation.

    \item ZS-TOD~\cite{Mosharrof2023ZeroShotGE} introduced a zero-shot generalizable E2E TOD model that incorporates domain schema and dialog annotations to generate dialog responses.

    \item Combiner~\cite{Chen2022KETODKT} introduced a pipeline of a TOD model followed by a response generation model. The pipeline TOD model is built using the SimpleTOD model and generates the belief states and actions, which are fed into a GPT-2-based model for generating the system response.
    
    \item SimpleTODPlus~\cite{Chen2022KETODKT} introduced a TOD model based on SimpleTOD.

    \item MinTL~\cite{Lin2020MinTLMT} introduced Levenshtein belief spans (Lev) that model the difference between the current and previous dialog state. A response is generated by conditioning on the dialog context and knowledge base results. The BiToD leaderboard provides implementation of MinTL using mbart, mt5. 

\end{itemize}

\vspace{-9pt}
\section{Results}
\vspace{-3pt}

\begin{table*}[!t]
    \centering
    \begin{adjustbox}{max width=\textwidth}
        \begin{tabular}{|c |c|c | c|R c|c|R c|c|R c|c|R c|c|c|}
            \hline
    \multirow{2}{*}{\textbf{Dataset}} & \multirow{2}{*}{\textbf{Model}} & \multicolumn{3}{R}{\textbf{API Invoke Accuracy}} & \multicolumn{3}{R}{\textbf{API Method Accuracy}} & \multicolumn{3}{R}{\textbf{Param Names Accuracy}} & \multicolumn{3}{R}{\textbf{Param Values Accuracy}} & \multicolumn{3}{c|}{\textbf{Full API Accuracy}} \\ \cline{3-17}
    & & \textbf{all}  & \textbf{seen} & \textbf{unseen} & \textbf{all}  & \textbf{seen} & \textbf{unseen} & \textbf{all}  & \textbf{seen} & \textbf{unseen} & \textbf{all}  & \textbf{seen} & \textbf{unseen} & \textbf{all}  & \textbf{seen} & \textbf{unseen} \\ \hline
    \multirow{4}{*}{SGD} & {\gptt} &68.42  & 69.16 & 68.01 
    & 66.09 & 66.98 & 65.38 
    & 67.24 & 64.84 & 68.38 
    & 61.37 & 60.56 & 61.72 
    & 33.32 & 32.63 & 28.52 
    \\ 
    & \gpt & 86.29  & 94.24 & 81.39 
    & 75.16 & 91.98 &67.64 
    & 58.21 & 87.42 &47.38 
    & 56.21 & 86.57 &44.87
    & 34.62 & 77.96 &17.82 
    \\ 
    
    & \flan & \textbf{98.65}  & \underline{99.61} & \textbf{98.24} 
    & \textbf{96.84} & \underline{99.61} & \textbf{95.23} 
    & \underline{79.71} &\underline{96.97} & \underline{74.58} 
    & \underline{76.03} &\underline{94.04} & \underline{70.43}
    & \underline{56.64} &\underline{89.49} & \underline{41.96} 
    \\ 
 
    & \opt & \underline{97.02}  & \textbf{99.84} & \underline{95.81} 
    & \underline{94.74} & \textbf{99.84} & \underline{92.13} 
    & \textbf{90.41} & \textbf{98.19} & \textbf{87.86} 
    & \textbf{87.70} & \textbf{97.91} & \textbf{84.02}     & \textbf{63.81} & \textbf{90.97} & \textbf{51.98} 
    \\ 
    \hline

    \multirow{4}{*}{KETOD} & {\gptt} &64.57 & 65.10 & 63.22
    & 62.12 & 62.08 & 62.99
    & 63.47 & 61.17 & 67.11
    & 57.64 & 56.03 & 60.30
    & 29.59 & 30.54 & 23.45
    \\ 
    & \gpt &78.83  & 80.87 & 73.56 
    & 71.48 & 78.36 & 62.53 
    & 57.02 & 72.98 & 39.60 
    & 55.46 & 71.83 & 37.73 
    & 36.75 & 59.56 & 10.80 
    \\ 
    & \flan &\textbf{90.45}  & \textbf{89.93} & \textbf{90.34} 
    & \textbf{89.07} & \textbf{89.93} & \textbf{86.67} 
    & \underline{78.62} & \textbf{88.92} & \underline{67.93} 
    & \underline{76.52} & \textbf{87.94} & \underline{65.28} 
    & \textbf{59.42} & \textbf{83.72} & \underline{35.86} 
    \\ 

     & \opt &\underline{82.22}  & \underline{81.38} & \underline{84.14} 
    & \underline{78.45} & \underline{80.87} & \underline{77.70} 
    & \textbf{79.50} & \underline{83.85} & \textbf{78.94} 
    & \textbf{77.43} & \underline{83.46} & \textbf{75.87} 
    & \underline{57.35} & \underline{77.18} & \textbf{39.54} 
    \\  
  \hline

    \multirow{4}{*}{BiToD} & {\gptt} &44.59& 44.59 & - 
    & 08.31& 08.31 & - 
    & 04.53& 04.53 & - 
    & 04.15& 04.15 & - 
    & 02.36& 02.36 & - 
    \\ 
    & \gpt &93.46  & 93.46 & - 
    & 87.34 & 87.34 & - 
    & 86.29 & 86.29 & - 
    & 86.10 & 86.10 & - 
    & 83.19 & 83.19 & - 
    \\ 
    & \flan &\textbf{98.48}  & \textbf{98.48} & - 
    & \textbf{93.07} & \textbf{93.07} & - 
    & \underline{90.76} & \underline{90.76} & - 
    & \underline{85.47} & \underline{85.47} & - 
    & \underline{73.79} & \underline{73.79} & - 
    \\ 
     
     & \opt &\underline{97.36}& \underline{97.36} & - 
    & \underline{92.70}& \underline{92.70} & - 
    & \textbf{92.32}& \textbf{92.32} & - 
    & \textbf{92.17}& \textbf{92.17} & - 
    & \textbf{91.11}& \textbf{91.11} & - 
    \\ 
 \hline
        \end{tabular}
    \end{adjustbox}
    \vspace{-6pt}
    \caption{Detailed results for API calls.}
    \label{tab:api_results}
    \vspace{-12pt}
\end{table*}




\vspace{-3pt}
\subsection{Response Generation Task}
\label{sec:response_generation}
\vspace{-3pt}
\textbf{Comparison with OpenAI’s GPT Models.} Table~\ref{tab:response_results} compares the performance of all models in {\oursys} with OpenAI’s {\gptt} for overall response generation and sub-tasks inform and request. 
The results show that all models employed in {\oursys} consistently outperform {\gptt} by large margins across all response categories on all three datasets.
The superior performance of {\oursys} suggests that instruction fine-tuning is crucial for handling TOD systems effectively.
Prompt-based approaches, while effective in many tasks, do not seem to handle the intricacies of TOD systems as multi-task instruction fine-tuned models do.
It is important to note that the template for {\oursys} is simple, requiring minimal instructions and no examples.
In contrast, the template for {\gptt} is more complex. 
Crafting an effective prompt involved an iterative process, where multiple templates were tested and refined to achieve optimal performance.
Figure~\ref{fig:finetuning_template} and Figure~\ref{fig:chatgpt_template} in Appendix~\ref{sec:templates} shows the prompt templates for {\oursys} and {\gptt}. 

\noindent
\textbf{Performance Comparison Across NL-TOD Models.}
The {\opt} model consistently demonstrates superior performance compared to {\gpt} and {\flan} across all metrics and datasets. 
From the results, we can see that a clear pattern emerges, where increasing the model size correlates with enhanced performance, underscoring the impact of model capacity on handling complex TOD tasks. 
Please note that {\opt} was fine-tuned using QLoRA, {\flan} with LoRA, and {\gpt} underwent full model fine-tuning.

Among the sub-tasks of response generation, we can see that the performance of all models is much lower for Request when compared to Inform. Consider a dialog example where a user asks, ``I am hungry and looking for a place to eat.''. There could be multiple potential pieces of information that the model could request, for example, \texttt{cuisine}, \texttt{city}, \texttt{price\_range}. 
However, in the dataset, if the system is expected to ask about \texttt{cuisine} but the model asks about \texttt{city}, it will be penalized, thus making the Request sub-task scores quite low.

\noindent
\textbf{Comparison with SOTA Approaches.} Table~\ref{tab:sgd_sota} compares the performance of {\opt} (best model in {\oursys}) with SOTA approaches.  
It is important to recall that these SOTA methods use turn-level annotations and mainly focus on response generation using supervised learning, while {\oursys} is optimized for response generation as well as making API calls for the zero-shot setting.
Existing approaches do not consider the API calls task, so we compare only for the response generation. 
We see that on the SGD and KETOD datasets,  {\opt} outperforms existing SOTA methods, with performance improvements of \textbf{53.15\%} and \textbf{129.06\%}, respectively. 
Despite using additional annotations, SOTA methods fall short compared to the simple approach of {\oursys}. 
This significant performance gain underscores the effectiveness of multi-task instruction fine-tuning. 
The BiToD dataset includes only seen domains, which makes it more suitable for supervised learning methods and less favorable for zero-shot learning approaches. 
Please note that {\oursys} is designed to excel in zero-shot scenarios, where it must generalize to new, unseen domains without additional training. 
Despite this, it still outperforms SOTA supervised learning methods by \textbf{7.67\%} in seen domains.

\noindent
\textbf{Qualitative Analysis of Generations.} 
We also conduct qualitative analysis of comparative response generations by different models including Open AI's GPT models.
The responses from different models for a sample dialog are presented in Appendix~\ref{sec:example_dialogs}.
We notice the differences between the response generation styles of {\oursys} and Open AI's GPT models. Open AI's GPT models tend to produce long and descriptive responses and it often repeats significant amount of information from previous turns, which can make the conversation less efficient and more cumbersome for the user.
In contrast, models in {\oursys} generate responses that succinct, closely mirroring the style of the gold standard responses. 
This behavior indicates that {\oursys}, after undergoing the fine-tuning, has effectively learned the concise response style from the gold standard and applies it consistently in its outputs. 

\begin{table*}[!t]
    \centering
    \small
    \begin{adjustbox}{max width=\textwidth}
        \begin{tabular}{|c|c|c|c|c|c|c|c|c|c|}
            \hline
            \multirow{2}{*}{\textbf{Dataset}}&\multirow{2}{*}{\textbf{Model}} & \textbf{Response} & \textbf{Inform}  & \textbf{Request} & \textbf{API} & \textbf{API} & \textbf{Param} & \textbf{Param} & \textbf{Full} 
            \\ 
            & & \textbf{BLEU-4} & \textbf{BLEU-4} & \textbf{BLEU-4}& \textbf{Invoke} & \textbf{Method} & \textbf{Names} & \textbf{Values} & \textbf{API} 
            \\ \hline
            \multirow{2}{*}{SGD} & {\gptt}& 0.0337 & 0.0847 & 0.0075 & 69.00 & 68.00 & 69.21 & 63.35 & 32.33
            \\ 
            &\gptf& 0.0440 & 0.0699 & 0.0072 & 42.33 & 42.33 & 44.16 & 42.02 & 25.67
            \\ 
             \hline
            \multirow{2}{*}{KETOD}&{\gptt}& 0.0435 & 0.0952 & 0.0047 & 66.67 & 64.33 & 65.74 & 60.21 & 36.00
            \\ 
            & \gptf& 0.0543 & 0.0929 & 0.0156 & 31.00 & 30.00 & 31.42 & 30.34 & 21.00
            \\ \hline
            \multirow{2}{*}{BiToD} & {\gptt}& 0.0704 & 0.2716 & 0.0008 & 42.00 & 09.00 & 04.71 & 04.54 & 02.00
            \\ 
            & \gptf& 0.0634 & 0.1949 & 0.0035 & 20.00 & 02.50 & 01.31 & 01.18 & 00.00
            \\ \hline
            \multicolumn{2}{|c|}{{\gptf} \% improvement} & \textbf{15.15} $\uparrow$ & \textbf{-16.04} $\downarrow$& \textbf{188.47} $\uparrow$ & \textbf{-48.18} $\downarrow$& \textbf{-54.45} $\downarrow$& \textbf{-53.53}$\downarrow$ & \textbf{-52.43} $\downarrow$&\textbf{-54.09} $\downarrow$ 
            \\ \hline
            
        \end{tabular}
    \end{adjustbox}
    \vspace{-6pt}
    \caption{Performance comparison of {\gptt} and {\gptf} for 1000 sample turns.}
    \label{tab:chatgpt_comparison}
    \vspace{-12pt}
\end{table*}






\vspace{-5pt}    
\subsection{API Call task}
\vspace{-1pt}
\noindent
\textbf{Comparison with OpenAI's GPT Models.} Table~\ref{tab:api_results} presents the results of the API call task, where {\flan} and {\opt} models in {\oursys} significantly outperform {\gptt} across all metrics. 
A key metric here is API invoke accuracy, which assesses the system's ability to correctly identify when an API call is needed. 
{\gptt} shows notably lower accuracy in this area compared to other models, underscoring the importance of fine-tuning and context learning for making API calls.
Interestingly, {\gptt} performs better on the API call task than on natural language response generation, even outperforming GPT-2 Medium on some metrics in unseen domains, such as full API accuracy on the SGD and KETOD datasets. 
This may be due to {\gptt}'s extensive exposure to programming code during pre-training, which aids in constructing API queries. 
However, its low API invoke accuracy suggests that while Open AI's GPT models can generate API queries effectively, it struggles with determining the correct timing for these calls within a dialog.

Table~\ref{tab:multi_api_mixed_results} in Appendix~\ref{sec:multi_dom_api} presents the API call results when the model must carry information between domains with different parameter names. Here, too, {\oursys} consistently outperforms {\gptt} across all metrics. 
Notably, {\gptt} shows relatively better performance compared to its results across all API calls.
This improved performance of {\gptt} may be attributed to its extensive pre-training, during which it likely encountered numerous examples of programming code.

\noindent
\textbf{Performance Comparison Across NL-TOD Models.} 
Among the models in {\oursys}, no single model consistently outperforms the others across all metrics. For example, {\opt} excels in unseen domains for full API call accuracy, a more challenging metric than correctly generating individual API components. {\opt} also generally performs better on the SGD dataset, largely due to its superior ability to accurately identify the parameter names required for API calls -- a crucial factor in this task.
Similar to response generation, performance tends to improve with larger model sizes, especially in full API call accuracy. 
However, the results on the KETOD dataset introduce some nuances: while {\flan} outperforms {\opt} on most metrics, it lags behind in unseen domains, particularly in parameter name accuracy and full API call accuracy.


\noindent
\textbf{Qualitative Analysis of API Calls.} 
We qualitatively analyze API call generations (see Appendix~\ref{sec:example_dialogs}) to compare how {\oursys} and Open AI's GPT models handle API calls.
Open AI's GPT models often makes API calls without first confirming with the user, leading to errors in the conversation, particularly around the API call itself.
This suggests that Open AI's GPT models struggles with the conversational nuances and timing necessary for effective API usage, resulting in mistimed calls.
In contrast, {\oursys} better understands the importance of confirming with the user before making an API call. 
{\oursys} reduces the errors observed with Open AI's GPT models, resulting in accurate results from API calls, which enhances the overall reliability of the TOD system.


\vspace{-2pt}
\subsection{Comparison with {\gptf}}
\vspace{-2pt}
Table~\ref{tab:chatgpt_comparison} compares the performance of {\gptt} and {\gptf} across 1,000 sampled turns, including 700 system response turns and 300 API call turns. The results indicate that {\gptf} excels in response generation, while {\gptt} performs better in API call tasks. 
While one might argue that {\gptf}, being the superior model, should have been the focus of a full comparison, our findings suggest otherwise. 
Although {\gptf} excels in response generation and request sub-task, {\gptt}, with its significantly lower cost, is better suited for TOD systems where API calls and task completion are critical.

\vspace{-2pt}
\section{Conclusion}
\vspace{-2pt}
This work introduces a novel zero-shot generalizable task-oriented dialog (TOD) system, {\oursys}, which reduces reliance on manually annotated turn-level data. Such annotations are often costly, error-prone, and inconsistent. Our approach enables researchers and practitioners to utilize the vast amounts of available conversational data to train TOD systems, eliminating the need for extensive manual labeling.
We consider API calls a core task, as it is crucial for a TOD system to know when and how to interact with external data sources.
To demonstrate the possibility of training TOD systems effectively without turn-level annotations, we employed multi-task instruction fine-tuning and trained three LLMs of varying sizes.
To promote zero-shot generalization, we use domain schema as an additional conditioning variable.
In fine-tuning these models, we experimented with both full-model training and low-rank adaptation techniques. 
We conducted extensive experiments using three diverse TOD datasets, analyzing overall response and API generation capabilities, as well as performance on sub-tasks.
Our results show that {\oursys} outperforms OpenAI's GPT models and existing state-of-the-art (SOTA) approaches that rely on turn-level annotated dialogs.

\section{Limitations}

{\oursys} has been developed by fine-tuning LLMs such as {\opt}, {\flan}, and {\gpt}. 
These LLMs require significant computational resource requirements to train, particularly {\opt}. 
Training and inference with these models can be expensive, limiting their practicality for deployment in resource-constrained environments. 

Experimental results show that while models perform well on seen domains, their ability to generalize to unseen domains is still a challenge. So the system may struggle with user queries that deviate from the expected task structure. 
The LLMs used in the system function as black boxes, making it challenging to interpret the reasoning behind their responses. 
This lack of transparency hinders the ability to diagnose and correct erroneous outputs, which is crucial in TOD systems where accuracy is critical. Furthermore, the models may inherit biases present in the training data, leading to biased or unfair responses in certain scenarios. 
Although efforts were made to mitigate this issue by fine-tuning using the dialog datasets, completely eliminating biases remains a challenging task. 
The reliance on pre-trained models introduces limitations related to the coverage of the pre-training data. If the pre-training data lacks specific domain knowledge, the TOD system may underperform in those domains.

The deployment of LLMs in TOD systems raises ethical and privacy concerns, particularly regarding the handling of sensitive user data. Ensuring that the system complies with privacy regulations and ethical standards is an ongoing challenge that requires continuous monitoring and updates. 
Similar to other AI technologies, there is a scope for potential misuse of our system. 
If {\oursys} is used with malicious intent or the model is fed inappropriate data, there is a risk of abuse. 
We would strongly advise to take necessary precautions and appropriate usage policies.

Addressing the limitations outlined above is crucial for advancing the effectiveness and reliability of TOD systems.
While the usage of pre-trained LLMs offers significant advantages, these models are not without their challenges. 
Increasing model interpretability, mitigating biases, and addressing ethical and societal concerns are essential steps toward creating more robust and responsible TOD systems. 

\bibliography{custom}

\appendix


\section{Templates for Instruction fine-tuning Open source Models and OpenAI's GPT Models}
\label{sec:templates}

\begin{figure*}
   \centering
   \includegraphics[width=0.95\linewidth]{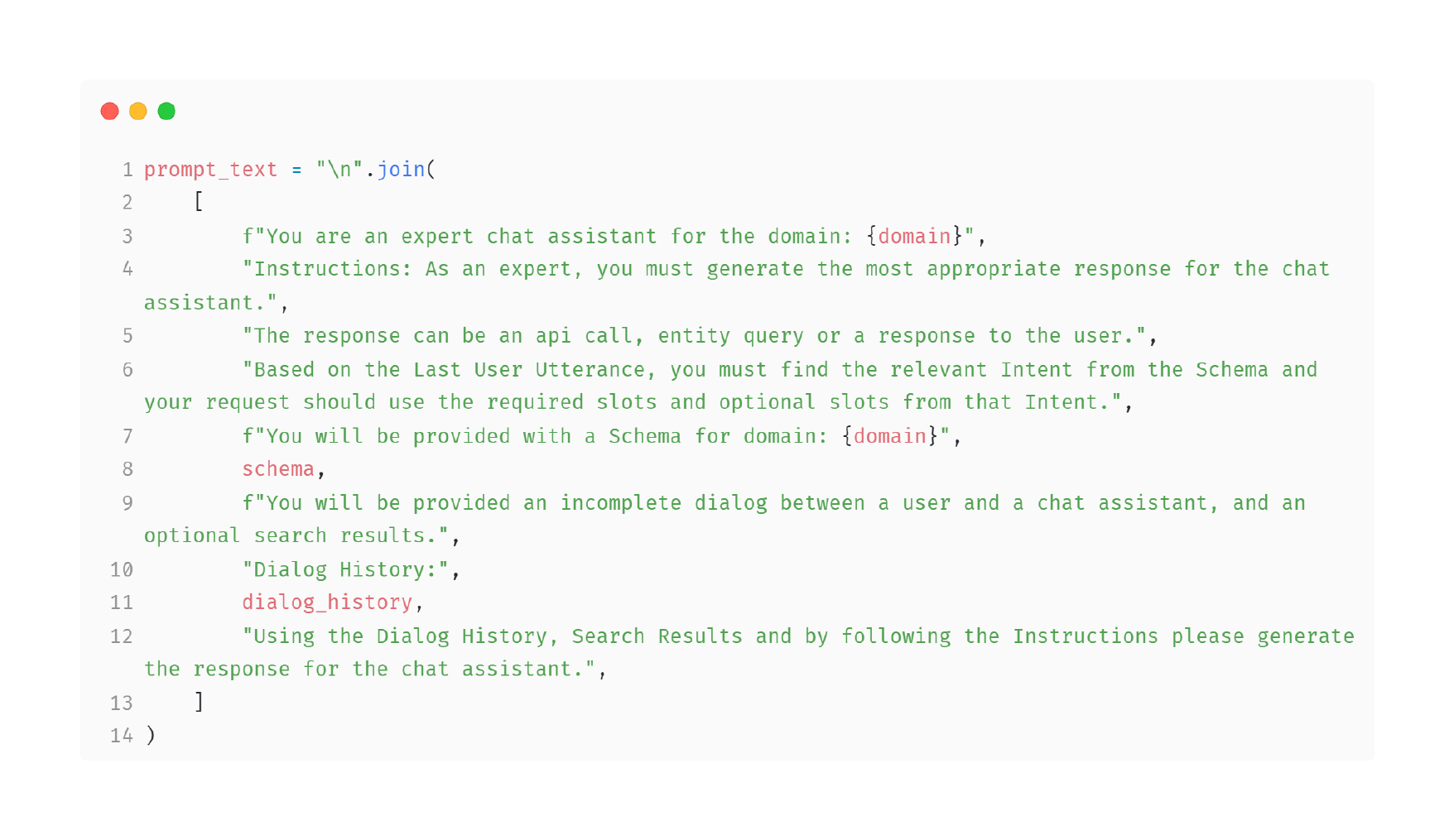}
   
   \caption{
       Template used to generate the prompt for fine-tuning {\oursys}.
   }
   
   \label{fig:finetuning_template}
\end{figure*}

Figure~\ref{fig:finetuning_template} shows the template used to process the inputs for {\oursys}. The template first informs about the domains involved in the dialog. Next, it provides task-specific instructions and provides the domain schema. The dialog history is provided and the model is instructed to generate a system response based on the dialog history, search results, and the task instructions.

\begin{figure*}
   \centering
   \includegraphics[width=0.95\linewidth]{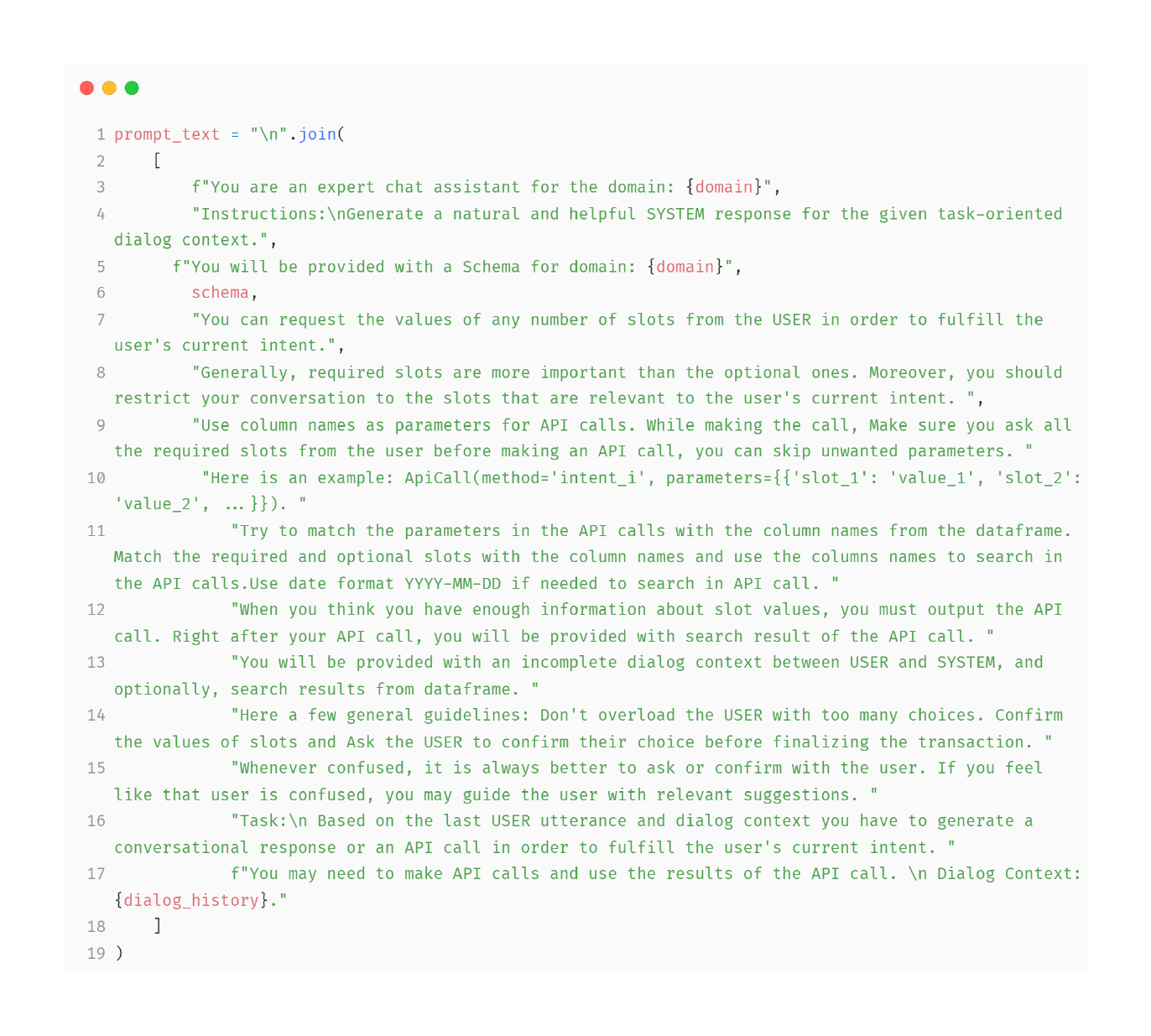}
   \caption{
       Template used to generate the prompt for OpenAI's GPT Models. 
   }
   
   \label{fig:chatgpt_template}
\end{figure*}

Figure~\ref{fig:chatgpt_template} shows the template used to process the inputs for OpenAI's GPT Models. Similar to the template of {\oursys}, information about the domains is provided first. Next task specific instructions are provided followed by the domain schemas. Next, comprehensive instructions are provided about the schema and how to make API calls. Detailed instructions on how to interact with a user are provided followed by instructions on how to use the information provided in the dialog context.
\section{Additional Results for Response Generation}
\label{sec:appendix_gleu_mixed}

Tables~\ref{tab:response_gleu_mixed},~\ref{tab:inform_gleu_mixed},~\ref{tab:seeking_information_gleu_mixed} presents the results for the system response and the sub-tasks for the response generation task. We also provide the GLEU metric and the mixed domain setting.
From Tables~\ref{tab:response_gleu_mixed},~\ref{tab:inform_gleu_mixed}, for the overall response and inform sub task, we can see that across all the datasets, the BLEU and GLEU metrics improve as the size of the model increases, with {\opt} producing the best results. However, in Table~\ref{tab:seeking_information_gleu_mixed} we can see a slight difference in the KETOD dataset, where the {\flan} model achieves the highest performance on the GLEU score on the mixed domains. For all other settings {\opt} produces the best performance. Another interesting observation in Tables~\ref{tab:response_gleu_mixed},~\ref{tab:inform_gleu_mixed},~\ref{tab:seeking_information_gleu_mixed} is that the second best performing model is {\gpt}. A possible explanation of this could be that the BiToD dataset is for supervised settings and is quite small when compared to SGD and KETOD, thus the {\gpt} model was able to effectively learn the patterns present in the dataset and achieved better results.

\section{Results to Carry Information from Domain One to Next in Multi-domain Dialogs}
\label{sec:multi_dom_api}

Table~\ref{tab:multi_api_mixed_results} presents the results for Carry Information from Domain One to Next in Multi-domain Dialogs. 
For this setting, we only have results on the SGD and KETOD datasets. We can see that for the key metric, Full API Accuracy, {\opt} achieves the best results across both datasets. {\opt} also achieves the highest scores for the Parameter Names Accuracy and Parameter Values Accuracy metrics as well. However, for the API Invoke Accuracy and API Method Accuracy, on some settings, {\flan} outperforms {\opt}.
Upon comparing the key metric, Full API Accuracy, of single domain results from Table~\ref{tab:api_results} with that of Multi-domain Dialogs, we can see that the performance of the models is better in the Multi-domain dialogs. A possible explanation of this could be that a bulk of the multi-domain dialogs contain common and generic domains like \texttt{Events}, \texttt{Restaurants}, and \texttt{Hotels}.
The dataset contains a lot of example dialogs from these domains, which in turn could have been the reason behind the improved performance. 

We can see that apart from the API Invoke Accuracy and API Method Accuracy, {\opt} is the model that produces the best results for the remaining metrics across both results. {\flan} is the second best performing model, and also achieves best results on the two metrics mentioned before. A notable observation on the key metric, Full Api Accuracy, is that the performance for mixed domain settings is in between that of the seen and unseen domain settings. This is the expected result, as mixed domains contain domains from both seen and unseen domains.

\section{Example Dialogs with Responses from All Models}
\label{sec:example_dialogs}


Table~\ref{tab:rest_dialog} shows an example dialog in the Restaurant domain and Table~\ref{tab:multi_domain_dialog} shows an example multi-domain dialog in the Buses and \texttt{RideShare} domain. The table contains the turn id, user utterance, gold response, {\gptt}, {\gptf}, {\gpt}, {\flan}, and {\opt} response. Text highlighted in red outlines the portions where the system response by a model is incorrect and green highlights the correct parts. Texts highlighted in orange indicate that the model is partially correct and is missing some information. Texts highlighted in yellow indicate that the system response seems like a valid response but is incorrect because the gold standard is expecting a different response. 

From the example dialogs, we can see that {\gptt} and {\gptf} has a common mistake of making an API Call without confirming with the user, whereas {\oursys} models do not make this mistake. Also, we can see that the responses of {\gptt} and {\gptf} are longer and more descriptive, whereas {\oursys} models produce succinct responses, similar to the gold standards. This observation indicates that {\oursys} was able to pick up the text generation style of the gold standard, which is an important aspect for TOD systems.

We can see that {\gpt} hallucinates and makes up information on its own, but {\flan} and {\opt} do not suffer from hallucination as much. For the API call task, all the models face similar problems of missing certain parameters and in the multi-domain setting, the models fail to adapt to the slots of the new domain. For response generation, on turns 1 in Table~\ref{tab:multi_domain_dialog}, we can see the request sub-task challenge as discussed in Section~\ref{sec:response_generation}, where the gold standard expects a certain slot but the models request a different slot. Another observation from turn 2 of that dialog is that the gold standard expects the model to request 3 slots, but the {\oursys} models are only asking about a single slot. A reason for this could be that there fewer turns where the gold standard expects multiple slots to be requested, so the models picked this pattern and asked about one slot.

On turn 4 from Table~\ref{tab:rest_dialog}, we can see that {\gpt} only mentioned that the restaurant was booked but did not mention the cuisine of the restaurant, whereas all other models did not make this mistake. Similar examples exist on a few other turns, where {\gpt} misses important information.

\begin{table*}[!t]
    \centering
    \small
    \begin{adjustbox}{max width=\textwidth}
        \begin{tabular}{|c|c|c|c|c| R c|c|c|c|}
            \hline
    \multirow{2}{*}{\textbf{Dataset}}&\multirow{2}{*}{\textbf{Model}} & \multicolumn{4}{R}{\textbf{Overall Response (BLEU-4)}} & \multicolumn{4}{c|}{\textbf{Overall Response (GLEU-4)}}  \\ \cline{3-10}
    & & \textbf{all}  & \textbf{seen} & \textbf{unseen} & \textbf{mixed} & \textbf{all}  & \textbf{seen} & \textbf{unseen} & \textbf{mixed} \\ \hline
    \multirow{4}{*}{SGD}&{\gptt} & 0.0397& 0.0422 & 0.0363 & 0.043
    &  0.0555& 0.0570 & 0.0522 & 0.0589
    \\ 
    &\gpt& 0.2276& 0.2423 & 0.2095 & 0.2436
    &  0.2240& 0.2430 & 0.2060 & 0.2395
    \\ 
    &\flan& \underline{0.2680}& \underline{0.2980} & \underline{0.2500} & \underline{0.2790}
    &  \underline{0.2610}& \underline{0.2960} & \underline{0.2440} & \underline{0.2721}
    \\ 
    &\opt& \textbf{0.3063}& \textbf{0.3311} & \textbf{0.2896} & \textbf{0.3174}
    &  \textbf{0.2996}& \textbf{0.3210} & \textbf{0.2841} & \textbf{0.3115} 
 \\  \hline
    
    \multirow{4}{*}{KETOD}&{\gptt} &0.0397& 0.0412 & 0.0334 & 0.0438
    & 0.0554& 0.0563 & 0.0494 & 0.0601
    \\ 
    &\gpt&0.1509& 0.1756 & 0.1224 & 0.1507
    & 0.1552& 0.1751 & 0.1303 & 0.1568
    \\ 
    &\flan& \underline{0.2130}  & \underline{0.2503} & \underline{0.1794} & \underline{0.2040}
    & \underline{0.2091} & \underline{0.2408} & \underline{0.1825} & \underline{0.2000}
    \\ 
    &\opt&\textbf{0.2680}  & \textbf{0.2994} & \textbf{0.2371} & \textbf{0.2559}
    & \textbf{0.2534} & \textbf{0.2860} & \textbf{0.2248} & \textbf{0.2446}
 \\  \hline

        \multirow{4}{*}{BiToD}&{\gptt} &0.0646  & 0.0646 & - & -
    & 0.0760 & 0.0760 & - & -
    \\ 
    &\gpt&\underline{0.3466}  & \underline{0.3466} & - & -
    & \underline{0.3016} & \underline{0.3016} & - & -
    \\ 
    &\flan& 0.3159  & 0.3159 & - & -
    & 0.2640 & 0.2640 & - & -
    \\ 
    &\opt&\textbf{0.4143}& \textbf{0.4143} & - & -
    & \textbf{0.3803}& \textbf{0.3803} & - & -
 \\  \hline
    
        \end{tabular}
    \end{adjustbox}
    \caption{Overall Response Generation: BLEU \& GLEU scores for system generations with additional mixed domain setting.}
    \label{tab:response_gleu_mixed}
\end{table*}

\begin{table*}[!t]
    \centering
    \small
    \begin{adjustbox}{max width=\textwidth}
        \begin{tabular}{|c|c|c|c|c| R c|c|c|c|}
            \hline
    \multirow{2}{*}{\textbf{Dataset}}&\multirow{2}{*}{\textbf{Model}} & \multicolumn{4}{R}{\textbf{Inform (BLEU-4)}} & \multicolumn{4}{c|}{\textbf{Inform (GLEU-4)}}  \\ \cline{3-10}
    & & \textbf{all}  & \textbf{seen} & \textbf{unseen} & \textbf{mixed} & \textbf{all}  & \textbf{seen} & \textbf{unseen} & \textbf{mixed} \\ \hline
    \multirow{4}{*}{SGD}&{\gptt} &0.1047& 0.0998 & 0.1012 & 0.1094
    & 0.1563& 0.1505 & 0.1523 & 0.1616
    \\ 
    &\gpt& 0.2737& 0.3022& 0.2479 & 0.2913
    &  0.3644& 0.3880& 0.3421 & 0.3799
    \\ 
    &\flan& \underline{0.2991}& \underline{0.3064} & \underline{0.2872} & \underline{0.3099}
    &  \underline{0.3879}& \underline{0.3960} & \underline{0.3775} & \underline{0.3968}
    \\ 
    &\opt&\textbf{0.3334}& \textbf{0.3499} & \textbf{0.3190} & \textbf{0.3431}
    & \textbf{0.4199}& \textbf{0.4322} & \textbf{0.4092} & \textbf{0.4271}
 \\  \hline
    
    \multirow{4}{*}{KETOD}&{\gptt} &0.0943& 0.0801 & 0.0958 & 0.1108
    & 0.1439& 0.1337 & 0.1428 & 0.1578
    \\ 
    &\gpt& 0.1734& 0.1843 & 0.1498 & 0.1817
    &  0.2565& 0.2717 & 0.2336 & 0.2587
    \\ 
    &\flan&\underline{0.2311}& \underline{0.2435} & \underline{0.2120} & \underline{0.2335}
    & \underline{0.3100}& \underline{0.3243} & \underline{0.2897} & \underline{0.3112}
    \\
    &\opt & \textbf{0.2968} & \textbf{0.2981} & \textbf{0.3034} & \textbf{0.2889}
    &  \textbf{0.3726}& \textbf{0.3776} & \textbf{0.3746} & \textbf{0.3646} 
 \\  \hline

        \multirow{4}{*}{BiToD}&{\gptt}& 0.2589& 0.2589 & - &-
    &  0.3092& 0.3092 & - &-
    \\ 
    &\gpt& \underline{0.5226}& \underline{0.5226} & - & -
    &  \underline{0.5682}& \underline{0.5682} & - & -
    \\ 
    &\flan& 0.4795& 0.4795 & - & -
    &  0.5389& 0.5389 & - & -
    \\ 
    &\opt& \textbf{0.5248}& \textbf{0.5248} & - & -
    &  \textbf{0.5900}& \textbf{0.5900} & - & -
 \\  \hline
    
        \end{tabular}
    \end{adjustbox}
    \caption{BLEU \& GLEU scores for inform sub-task of response generation task, with additional mixed domain setting.}
    \label{tab:inform_gleu_mixed}
\end{table*}

\begin{table*}[!t]
    \centering
    \small
    \begin{adjustbox}{max width=\textwidth}
        \begin{tabular}{|c|c|c|c|c| R c|c|c|c|}
            \hline
    \multirow{2}{*}{\textbf{Dataset}}&\multirow{2}{*}{\textbf{Model}} & \multicolumn{4}{R}{\textbf{Request (BLEU-4)}} & \multicolumn{4}{c|}{\textbf{Request (GLEU-4)}}  \\ \cline{3-10}
    & & \textbf{all}  & \textbf{seen} & \textbf{unseen} & \textbf{mixed} & \textbf{all}  & \textbf{seen} & \textbf{unseen} & \textbf{mixed} \\ \hline
    \multirow{4}{*}{SGD}& {\gptt} &0.0057 & 0.0046 & 0.0056 & 0.0060
    & 0.0337 & 0.0299 & 0.0343 & 0.0343
    \\ 
    &\gpt& 0.0616& 0.0659 & 0.0630 & 0.0588
    &  0.1387& 0.1453 & 0.1403 & 0.1353
    \\ 
    &\flan&\underline{0.0657}& \underline{0.0742} & \underline{0.0698} & \underline{0.0582}
    & \underline{0.1552}& \underline{0.1687} & \underline{0.1586} & \underline{0.1477}
    \\ 
    &\opt& \textbf{0.0836}& \textbf{0.0928} & \textbf{0.0793} & \textbf{0.0868}
    &  \textbf{0.1767}& \textbf{0.1870} & \textbf{0.1721} & \textbf{0.1803}
 \\  \hline
    
    \multirow{4}{*}{KETOD}&{\gptt} &0.0033& 0.0036 & 0.0037 & 0.0026
    & 0.0305& 0.0285 & 0.0332 & 0.0300
    \\ 
    
    &\gpt& 0.0140& 0.0193& 0.0073 & 0.0148
    &  0.0649& 0.0750& 0.0561 & 0.0620
    \\ 
    
    &\flan&\underline{0.0587}& \underline{0.0781} & \underline{0.0430} & \underline{0.0521}
    & \underline{0.1440}& \underline{0.1707} & \underline{0.1256} & \textbf{0.1315}
    \\ 
    
    &\opt& \textbf{0.0640}& \textbf{0.0851}& \textbf{0.0509} & \textbf{0.0526}
    &  \textbf{0.1475}& \textbf{0.1742}& \textbf{0.1339} & \underline{0.1301}
    
 \\  \hline

        \multirow{4}{*}{BiToD}&{\gptt} &0.0011& 0.0011& - & -
    & 0.0219& 0.0219& - & -
    \\ 
    &\gpt& \underline{0.0323}& \underline{0.0323} & - & -
    &  \underline{0.1035}& \underline{0.1035} & - & -
    \\ 
    &\flan&0.0256& 0.0256 & - & -
    & 0.0811& 0.0811 & - & -
    \\ 
    &\opt& \textbf{0.0476}& \textbf{0.0476} & - & -
    &  \textbf{0.1260}& \textbf{0.1260} & - & -
    
 \\  \hline
    
        \end{tabular}
    \end{adjustbox}
    \caption{BLEU \& GLEU scores for request sub-task of response generation task, with additional mixed domain setting.}
    \label{tab:seeking_information_gleu_mixed}
\end{table*}

\begin{table*}[!t]
    \centering
    \begin{adjustbox}{max width=\textwidth}
        \begin{tabular}{|c|c|c|c|c|c|c|c|c|c|c|c|c|c|}
            \hline
            \multirow{2}{*}{\textbf{Metric}} & \multirow{2}{*}{\textbf{Model}} & \multicolumn{4}{c|}{\textbf{SGD}} & \multicolumn{4}{c|}{\textbf{KETOD}} & \multicolumn{4}{c|}{\textbf{BiToD}} \\ \cline{3-14}
            & & \textbf{all} & \textbf{seen} & \textbf{unseen} & \textbf{mixed} & \textbf{all} & \textbf{seen} & \textbf{unseen} & \textbf{mixed} & \textbf{all} & \textbf{seen} & \textbf{unseen} & \textbf{mixed} \\ \hline

            \multirow{2}{*}{API} 
            & \gptt & 68.42 & 69.16 & 68.01 & 68.65 & 64.57 & 65.10 & 63.22 & 65.06 & 44.59 & 44.59 & - & - \\ 
            \multirow{2}{*}{Invoke} & \gpt & 90.51 & 97.66 & 86.21 & 93.11 & 78.83 & 80.87 & 73.56 & 80.75 & 93.46 & 93.46 & - & - \\ 
            \multirow{2}{*}{Accuracy} & \flan & \textbf{98.65} & \underline{99.61} & \textbf{98.24} & \textbf{98.84} & \textbf{90.45} & \textbf{89.93} & \textbf{90.34} & \textbf{91.09} & \textbf{98.48} & \textbf{98.48} & - & - \\ 
            & \opt & \underline{97.02} & \textbf{99.84} & \underline{95.81} & \underline{97.59} & \underline{82.22} & \underline{81.38} & \underline{84.14} & \underline{81.64} & \underline{97.36} & \underline{97.36} & - & - \\ 

            \hline
            \multirow{2}{*}{API} 
            & \gptt & 66.09 & 66.98 & 65.38 & 66.57 & 62.12 & 62.08 & 62.99 & 61.50 & 08.31 & 08.31 & - & - \\ 
            \multirow{2}{*}{Method} & \gpt & 78.62 & 96.26 & 71.42 & 81.80 & 71.48 & 78.36 & 62.53 & 71.12 & 87.34 & 87.34 & - & - \\ 
            \multirow{2}{*}{Accuracy} & \flan & \textbf{96.84} & \underline{99.61} & \textbf{95.23} & \textbf{97.81} & \textbf{89.07} & \textbf{89.93} & \textbf{86.67} & \textbf{90.02} & \textbf{93.07} & \textbf{93.07} & - & - \\ 
            & \opt & \underline{94.74} & \textbf{99.84} & \underline{92.13} & \underline{96.17} & \underline{78.45} & \underline{80.87} & \underline{77.70} & \underline{76.47} & \underline{92.70} & \underline{92.70} & - & - \\ 

            \hline
            \multirow{2}{*}{Parameter} 
            & \gptt & 67.24 & 64.84 & 68.38 & 66.65 & 63.47 & 61.17 & 67.11 & 63.08 & 04.53 & 04.53 & - & - \\ 
            \multirow{2}{*}{Name} & \gpt & 60.65 & 92.07 & 49.52 & 64.69 & 57.02 & 72.98 & 39.60 & 53.57 & 86.29 & 86.29 & - & - \\ 
            \multirow{2}{*}{Accuracy} & \flan & \underline{79.71} & \underline{96.97} & \underline{74.58} & \underline{86.68} & \underline{78.62} & \textbf{88.92} & \underline{67.93} & \textbf{75.97} & \underline{90.76} & \underline{90.76} & - & - \\ 
            & \opt & \textbf{90.41} & \textbf{98.19} & \textbf{87.86} & \textbf{91.21} & \textbf{79.50} & \underline{83.85} & \textbf{78.94} & \underline{75.30} & \textbf{92.32} & \textbf{92.32} & - & - \\ 

            \hline
            \multirow{2}{*}{Parameter} 
            & \gptt & 61.37 & 60.56 & 61.72 & 61.21 & 57.64 & 56.03 & 60.30 & 57.28 & 04.15 & 04.15 & - & - \\ 
            \multirow{2}{*}{Values} & \gpt & 58.64 & 91.24 & 46.98 & 62.95 & 55.46 & 71.83 & 37.73 & 51.83 & 86.10 & 86.10 & - & - \\ 
            \multirow{2}{*}{Accuracy} & \flan & \underline{76.03} & \underline{94.04} & \underline{70.43} & \underline{84.51} & \underline{76.52} & \textbf{87.94} & \underline{65.28} & \textbf{73.09} & \underline{85.47} & \underline{85.47} & - & - \\ 
            & \opt & \textbf{87.70} & \textbf{97.91} & \textbf{84.02} & \textbf{89.08} & \textbf{77.43} & \underline{83.46} & \textbf{75.87} & \underline{72.22} & \textbf{92.17} & \textbf{92.17} & - & - \\ 

            \hline
            \multirow{2}{*}{Full} 
            & \gptt & 33.32 & 32.63 & 28.52 & 38.06 & 29.59 & 30.54 & 23.45 & 33.33 & 02.36 & 02.36 & - & - \\ 
            \multirow{2}{*}{Api} & \gpt & 36.72 & 83.41 & 19.40 & 43.48 & 36.75 & 59.56 & 10.80 & 32.62 & 83.19 & 83.19 & - & - \\ 
            \multirow{2}{*}{Accuracy} & \flan & \underline{56.64} & \underline{89.49} & \underline{41.96} & \underline{63.79} & \textbf{59.42} & \textbf{83.72} & \underline{35.86} & \textbf{51.87} & \underline{73.79} & \underline{73.79} & - & - \\ 
            & \opt & \textbf{63.81} & \textbf{90.97} & \textbf{51.98} & \textbf{69.42} & \underline{57.35} & \underline{77.18} & \textbf{39.54} & \underline{50.09} & \textbf{91.11} & \textbf{91.11} & - & - \\

            \hline
        \end{tabular}
    \end{adjustbox}
    \vspace{-6pt}
    \caption{Detailed results for API calls across different metrics and datasets along with mixed domain.}
    \label{tab:api_results_mixed}
    \vspace{-12pt}
\end{table*}

\begin{table*}[!t]
    \centering
    \begin{adjustbox}{max width=\textwidth}
        \begin{tabular}{|c|c|c|c|c|R c|c|c |R c|c|c|c|}
            \hline
            \multirow{2}{*}{\textbf{Metric}} & \multirow{2}{*}{\textbf{Model}} & \multicolumn{4}{R}{\textbf{SGD}} & \multicolumn{4}{R}{\textbf{KETOD}} & \multicolumn{4}{c|}{\textbf{BiToD}} \\ \cline{3-14}
            & & \textbf{all} & \textbf{seen} & \textbf{unseen} & \textbf{mixed} & \textbf{all} & \textbf{seen} & \textbf{unseen}  & \textbf{mixed} & \textbf{all} & \textbf{seen} & \textbf{unseen}  & \textbf{mixed} \\ \hline
            
            \multirow{2}{*}{Api} 
            & \gptt & 68.34 & 73.42 & 67.77 & 68.12 & 62.05 & 61.05 & 66.00 & 61.20 & 29.47 & 29.47 & - & - \\ 
            \multirow{2}{*}{Invoke} & \gpt & 92.32 & 97.75 & 89.06 & 93.95 & 84.02 & 85.26 & 81.33 & 84.15 & \textbf{99.90} & \textbf{99.90} & - & - \\ 
            \multirow{2}{*}{Accuracy} & \flan & \textbf{98.80} & \underline{98.87} & \textbf{98.42} & \textbf{99.00} & \textbf{89.01} & \textbf{88.07} & \underline{90.00} & \textbf{89.34} 
            & 95.14 & 95.14 & - & -  \\ 
            & \opt & \underline{96.32} & \textbf{99.55} & \underline{94.98}  & \underline{96.87} & \underline{88.64} & \underline{87.72} & \textbf{92.67} & \underline{87.70} & \underline{99.31} & \underline{99.31} & - & - \\ 

            \hline
            \multirow{2}{*}{Api} 
            & \gptt & 65.94 & 72.07 & 64.80 & 66.01 & 60.05 & 58.60 & 66.67 & 58.47 & 12.23 & 12.23 & - & - \\ 
            \multirow{2}{*}{Method} & \gpt & 81.24 & 97.07 & 73.76 & 84.57 & 78.65 & 83.51 & 68.67 & 78.96 & 91.29 & 91.29 & - & - \\ 
            \multirow{2}{*}{Accuracy} & \flan & \textbf{96.94} & \underline{98.87} & \textbf{95.05}& \textbf{97.97} 
            & \textbf{87.02} & \textbf{88.07} & \underline{83.33} & \textbf{87.70} & \underline{91.53} & \underline{91.53} & - & - \\ 
            & \opt & \underline{93.87} & \textbf{99.55} & \underline{90.80} & \underline{95.33} & \underline{86.27} & \underline{87.72} & \textbf{86.00} & \underline{85.25} & \textbf{92.41} & \textbf{92.41} & - & - \\ 

            \hline
            \multirow{2}{*}{Parameter} 
            & \gptt & 65.85 & 68.99 & 64.65 & 66.31 & 60.07 & 58.10 & \underline{64.27} & 59.89 & 09.17 & 09.17 & - & - \\ 
            \multirow{2}{*}{Names} & \gpt & 58.23 & 95.61 & 44.97 & 63.01 & 61.22 & 79.28 & 35.37 & 57.76 & \underline{90.40} & \underline{90.40} & - & - \\ 
            \multirow{2}{*}{Accuracy} & \flan & \underline{77.31} & \underline{98.47} & \underline{72.96} & \underline{83.18} 
            & \underline{75.18} & \underline{87.11} & 62.96 & \underline{70.89} & 88.43 & 88.43 & - & - \\ 
            & \opt & \textbf{89.30} & \textbf{99.19} & \textbf{86.38} & \textbf{90.16} & \textbf{82.17} & \textbf{87.78} & \textbf{79.44} & \textbf{78.92} & \textbf{92.01} & \textbf{92.01} & - & -  \\ 

            \hline
            \multirow{2}{*}{Parameter} 
            & \gptt & 61.39 & 65.19 & 58.88 & 62.69 & 56.18 & 55.20 & 57.80 & 56.28 & 08.60 & 08.60 & - & - \\ 
            \multirow{2}{*}{Values} & \gpt & 56.46 & 94.52 & 42.59 & 61.60 & 59.44 & 77.53 & 33.22 & 56.09 & \underline{90.29} & \underline{90.29} & - & - \\ 
            \multirow{2}{*}{Accuracy} & \flan & \underline{73.77} & \underline{96.32} & \underline{68.36} & \underline{81.18}
            & \underline{73.16} & \underline{85.87} & \underline{59.72}& \underline{68.77} & 82.90 & 82.90 & - & - \\ 
            & \opt & \textbf{86.61} & \textbf{98.77} & \textbf{82.10} & \textbf{88.30}& \textbf{79.75} & \textbf{87.16} & \textbf{75.55} & \textbf{75.70} & \textbf{91.95} & \textbf{91.95} & - & - \\ 

            \hline
            \multirow{2}{*}{Full} 
            & \gptt & 38.62 & 40.77 & 33.66 & 41.83 & 34.08 & 36.84 & 26.67 & 34.97 & 03.55 & 03.55 & - & - \\ 
            \multirow{2}{*}{Api} & \gpt & 36.31 & 87.39 & 19.32 & 42.07 & 43.07 & 68.07 & 12.00 & 36.34 & \underline{88.14} & \underline{88.14} & - & - \\ 
            \multirow{2}{*}{Accuracy} & \flan & \underline{50.72} & \underline{92.12} & \underline{35.09} & \underline{56.39} 
            & \underline{56.30} & \underline{82.46} & \underline{30.00}& \underline{46.72} & 72.79 & 72.79 & -& -  \\ 
            & \opt & \textbf{61.13} & \textbf{96.40} & \textbf{46.91} & \textbf{66.85}& \textbf{60.55} & \textbf{83.16} & \textbf{36.67} & \textbf{52.73} & \textbf{91.38} & \textbf{91.38} & - & - \\ 

            \hline
        \end{tabular}
    \end{adjustbox}
    \caption{Results for carry information from domain one to next in multi-domain dialogs along with mixed domains.}
    \label{tab:multi_api_mixed_results}
\end{table*}

\onecolumn
\begin{longtblr}[
  caption = {Dialog from Restaurant domain in SGD dataset. Dialog id 1\_00001},
  label = {tab:rest_dialog},
]{
  colspec = {r|X[5]},
  rowhead = 0,
  hlines,
  row{even} = {},
  row{1} = {},
} 
            Turn & 1 \\ 
            User & Can you book a table for me at the Ancient Szechuan for the 11th of this month at 11:30 am?  \\ 
            Gold & In which city are you trying to book the table? \\ 
            \gptt & \color{BrickRed} Sure, I can help you with that. Just to confirm, you would like to make a reservation at Ancient Szechuan for the 11th of this month at 11:30 am, is that correct? \\ 
            \gptf &\color{ForestGreen}  Could you please confirm the location where you'd like to book the table at Ancient Szechuan? Additionally, how many seats will you need for the reservation?
 \\
            \gpt & \color{BrickRed} Sure, please confirm you'd like a table for 2 at Ancient Szechuan in San Jose at 11:30 am on March 11th.\\ 
            \flan & \color{ForestGreen} Sure, where are you located?  \\ 
            \opt & \color{ForestGreen} Sure, what city is that in? \\

            Turn & 2 \\ 
            User &  Can you book a table at the Butterfly restaurant in San Francisco?  \\ 
            Gold & Please confirm that you need a table for 2 at the Butterfly Restaurant in San Francisco at 11:30 am on March 11th. \\ 
            \gptt & \color{BrickRed} ApiCall(method=`ReserveRestaurant', parameters=`restaurant\_name': `Butterfly', `location': `San Francisco')\\ 
            \gptf &  \color{ForestGreen} Could you please confirm the date and time you would like to reserve the table at Butterfly Restaurant in San Francisco? Also, how many seats will you need?
\\
            \gpt & \color{ForestGreen} Please confirm that you want to book a table for 2 at the Butterfly restaurant in San Francisco. The reservation is for March 11th at 11:30 am.
 \\ 
            \flan & \color{ForestGreen} Please confirm the following details: Booking a table for 2 at Butterfly Restaurant in San Francisco on March 11th at 11:30 am. \\ 
            \opt & \color{ForestGreen} Please confirm that you want me to book a table for 2 at the Butterfly Restaurant in San Francisco on March 11th at 11:30 am. \\ 
            
            Turn & 3 \\ 
            Gold & ApiCall(method=`ReserveRestaurant', parameters= `date': `2019-03-11', `location': `San Francisco', `number\_of\_seats': `2',`restaurant\_name': `Butterfly Restaurant', `time': `11:30' )\\ 
            \gptt & ApiCall(\color{BrickRed} method=`FindRestaurants', parameters=`category': `All', `location': `San Francisco') \\ 
            \gptf & \color{BrickRed} Great, the reservation is nearly complete. As for the cuisine, the Butterfly Restaurant mainly offers modern American cuisine with a focus on seasonal ingredients. Would you like to proceed with the booking for March 11th at 11:30 am at the Butterfly Restaurant?
 \\
            \gpt & ApiCall(method=`ReserveRestaurant', \color{BrickRed}parameters=`city': `San Francisco', \color{ForestGreen}`date': `2019-03-11', \color{BrickRed}`party\_name': `Luxury Szechuan', \color{ForestGreen}`time': `11:30')

\\ 
            \flan & \color{ForestGreen} ApiCall(method=`ReserveRestaurant', parameters= `location': `San Francisco', `date': `2019-03-11', \color{BrickRed} `party\_size': `2', \color{ForestGreen}`restaurant\_name': `Butterfly Restaurant', `time': `11:30' )\  \\ 
            \opt & \color{Orange} ApiCall(method=`ReserveRestaurant', parameters=`location': `San Francisco', `date': `2019-03-11', `restaurant\_name': `Butterfly Restaurant', `time': `11:30') \\

            Turn & 4 \\ 
            User & That's perfect. What cuisine do they offer?  \\ 
            Gold & \color{ForestGreen} Your table has been booked successfully and they serve Asian cuisine. \\ 
            \gptt & \color{ForestGreen}The Butterfly Restaurant in San Francisco offers Asian cuisine. \color{BrickRed}Would you like me to proceed with booking a table for you at the Butterfly Restaurant in San Francisco for 2 people on March 11th at 11:30 am?\\ 
            \gptf & \color{ForestGreen} Your reservation at Butterfly Restaurant in San Francisco is confirmed for 2 people on March 11th at 11:30 am. The restaurant is located at 33 The Embarcadero and specializes in Asian cuisine. They also offer vegetarian options and have a moderate price range. Is there anything else I can assist you with today?
  \\
            \gpt & \color{Orange} Your table is booked.
\\
            \flan & \color{ForestGreen}Your reservation has been made. They serve Asian food. \\ 
            \opt & \color{ForestGreen} I've successfully booked your table. They offer Asian cuisine. \\

            Turn & 5 \\ 
            User & Can you get me the user rating of the restaurant? Is it expensive? \\ 
            Gold & Their prices are moderate and they have a user rating of 4.0. \\ 
            \gptt & \color{ForestGreen}The user rating of the Butterfly Restaurant is 4.00. It is considered to be in the moderate price range.\\ 
            \gptf & \color{ForestGreen} The Butterfly Restaurant in San Francisco has a user rating of 4.00 out of 5, and it falls within a moderate price range. Is there anything else you would like to know or another way I can assist you today?
\\
            \gpt & \color{ForestGreen}The restaurant is rated 4.0. The price range is moderate.
 \\
            \flan & \color{ForestGreen}The rating is 4.0 and the price range is moderate. \\ 
            \opt & \color{ForestGreen}The price range is moderate and the rating is 4.0. \\

            Turn & 6 \\ 
            User & Thanks a bunch! \\ 
            Gold & Is there anything else I can do for you? \\ 
            \gptt & \color{ForestGreen} You're welcome! If you have any more questions or need further assistance, feel free to ask. Enjoy your dining experience at Butterfly Restaurant!\\ 
            \gptf & \color{ForestGreen} You're welcome! If you need any more assistance or another reservation, just let me know. Enjoy your meal at Butterfly Restaurant!
 \\
            \gpt & \color{ForestGreen} Is there anything else I can help you with?\\
            \flan & \color{ForestGreen} Can I help you with anything else? \\ 
            \opt & \color{ForestGreen} Is there anything else I can do for you? \\

            Turn & 7 \\ 
            User & No. Thank you. That's all I need. \\ 
            Gold & Have a great day! \\ 
            \gptt & \color{ForestGreen}You're welcome! If you need any more assistance in the future, feel free to reach out. Have a great day!\\ 
            \gptf & \color{ForestGreen} You're welcome! If you need any more assistance in the future, feel free to ask. Have a wonderful day!
\\
            \gpt & \color{ForestGreen}Have a great day!
\\
            \flan & \color{ForestGreen}Have a great day! \\ 
            \opt & \color{ForestGreen}My pleasure. Have a great day! \\


\end{longtblr}

\twocolumn

\onecolumn
\begin{longtblr}[
  caption = {Multi domain dialog example from the Buses and RideShare domain in SGD dataset. Dialog id 12\_00006},
  label = {tab:multi_domain_dialog},
]{
  colspec = {r|X[5]},
  rowhead = 0,
  hlines,
  row{even} = {},
  row{1} = {},
} 
    Turn & 1 \\ 
    User & I need bus tickets. \\ 
    Gold & How many tickets? \\ 
    \gptt & \color{myyellow}Sure, I can help you with booking bus tickets. Can you please provide me with the following information: from which city, to which city, and on what date do you need the bus tickets? \\ 
    \gptf & \color{ForestGreen} Great, I can help you with that! Could you please provide me with the departure city, destination city, and the date you plan to travel? Additionally, it would be helpful to know how many passengers will be traveling and if you have a preferred bus category.
 \\
    \gpt & \color{myyellow}What time do you want to leave?
\\
    \flan & \color{myyellow} Where are you going? \\ 
    \opt & \color{myyellow} Where are you departing from? \\

    Turn & 2 \\ 
    User & For three people. \\
    Gold & When would you like to leave and at what time? Where are you heading to? \\   
    \gptt & \color{ForestGreen}Sure! I will need some more information to help you with the bus tickets. Can you please provide me with the departure city, destination city, and the date of travel?\\ 
    \gptf & \color{ForestGreen} Great! For which cities and what is the departure date for the bus tickets you need for three people?
\\
    \gpt & \color{orange}What time do you want to leave?
\\
    \flan & \color{orange} Where are you going? \\ 
    \opt & \color{orange} Where are you departing from? \\
    
    Turn & 3 \\ 
    User & I want to leave for Anaheim on the 11th of March at afternoon 3:30. \\ 
    Gold & Where would you be leaving from?  \\ 
    \gptt & \color{BrickRed} ApiCall(method=`FindBus', parameters={`from\_city': `current\_city', `to\_city': `Anaheim', `departure\_date': `2023-03-11', `num\_passengers': 3}) \\ 
    \gptf & \color{ForestGreen} Could you please confirm the city you'll be departing from to travel to Anaheim?
 \\
    \gpt & \color{BrickRed}You want 3 tickets on the bus leaving at 3:30 pm on March 11th at Anaheim.
\\
    \flan & \color{ForestGreen}Where are you leaving from? \\ 
    \opt & \color{ForestGreen} Where are you leaving from? \\
    
    Turn & 4 \\ 
    User & I would be leaving LA.  \\ 
    Gold & You would like tickets for 3 on a bus leaving Los Angeles for Anaheim on March 11th at 3:30 pm with no extra luggage. \\ 
    \gptt & \color{BrickRed} System: ApiCall(method=`BuyBusTicket', parameters={`from\_city': `LA', `to\_city': `Anaheim', `departure\_date': `2023-03-11', `departure\_time': `15:30', `num\_passengers': `3'}) \\ 
    \gptf & \color{BrickRed} ApiCall(method=`BuyBusTicket', parameters=\{`from\_city': `LA', `to\_city': `Anaheim', `departure\_date': `2023-03-11', `departure\_time': `15:30', `num\_passengers': `3'\})
 \\
    \gpt & \color{orange} You would like 3 tickets on a bus leaving at 3:30 pm on March 11th at 3:30 pm on the bus leaving from Los Angeles.
\\
    \flan & \color{ForestGreen} Please confirm: 3 tickets for the bus leaving Los Angeles on March 11th at 3:30 pm to Anaheim. \\ 
    \opt & \color{ForestGreen} You want 3 tickets for a bus from Los Angeles to Anaheim on March 11th at 3:30 pm? \\

    Turn & 5 \\ 
    Gold & ApiCall(method=`BuyBusTicket', parameters= `additional\_luggage': `False', `departure\_date': `2019-03-11', `departure\_time': `15:30', `from\_city': `Los Angeles', `num\_passengers': `3', `to\_city': `Anaheim' ) \\ 
    \gptt & \color{BrickRed} In order to get accurate information on the number of stops and the specific station the bus will be leaving from, it is best to contact the bus company directly. Would you like me to assist you in finding contact information for the bus company or would you like to handle it on your own? \\ 
    \gptf & \color{BrickRed}It looks like you're ready to purchase the bus tickets. However, I'll need to check the details of the bus trip including the number of stops and departure station. I'll proceed with that now.
\color{ForestGreen}ApiCall(method=`BuyBusTicket', parameters=\{`from\_city': `Los Angeles', `to\_city': `Anaheim', `departure\_date': `2023-03-11', `departure\_time': `15:30', `num\_passengers': 3\})
 \\
    \gpt & \color{orange}ApiCall(method=`BuyBusTicket', parameters=`departure\_date': `2019-03-11', `departure\_time': `15:30', `destination': `Anaheim', \color{BrickRed}`origin': `Los Angeles') \\
    \flan & \color{orange} ApiCall(method=`BuyBusTicket', parameters=`departure\_date': `2019-03-11', `departure\_time': `15:30', \color{ForestGreen}`to\_city': `Anaheim', `number\_of\_passengers': `3' ) \\ 
    \opt & \color{orange}ApiCall(method='BuyBusTicket', parameters={'from\_city': 'Los Angeles', 'departure\_date': '2019-03-11', 'departure\_time': '15:30', 'to\_city': 'Anaheim', 'num\_passengers': '3'}) \\
    
    Turn & 6 \\ 
    User & That would be fine. How many stops are there and which station would it be leaving from?  \\ 
    Gold & It is a direct bus and it would be leaving North Hollywood Station. I have bought the tickets. \\ 
    \gptt & \color{BrickRed} In order to get accurate information on the number of stops and the specific station the bus will be leaving from, it is best to contact the bus company directly. Would you like me to assist you in finding contact information for the bus company or would you like to handle it on your own? \\ 
    \gptf & \color{orange} The bus leaves from North Hollywood Station and arrives at Anaheim Bus Station. It's a direct bus with no stops. The total price for all three tickets will be \$39. Would you like to proceed with the purchase?
 \\
    \gpt & \color{ForestGreen}I have bought the tickets. There is a direct bus leaving at 3:30 pm and there are 0 stops.
 \\
    \flan & \color{ForestGreen} Your tickets have been confirmed. The bus will be leaving from North Hollywood Station and there are \color{BrickRed}10 stops. \\ 
    \opt & \color{ForestGreen}I have confirmed the tickets. There are only \color{BrickRed}1 stop and \color{ForestGreen}it leaves from North Hollywood Station.\\

    Turn & 7 \\ 
    User & I would need a nice SUV rental car in Anaheim on the 12th of this month. \\ 
    Gold & What time would you like to pick up the car and till when would you like it? \\ 
    \gptt & \color{BrickRed} The bus you're looking for is a direct route from Los Angeles to Anaheim on March 11th at 3:30 pm. The bus will be leaving from North Hollywood Station. Would you like to proceed with booking these tickets? \\ 
    \gptf & \color{ForestGreen} In order to assist you with the SUV rental in Anaheim, could you please provide the pick-up and drop-off times, as well as let me know if you'd like to add insurance to your booking?
\\
    \gpt & \color{BrickRed}You would like to pick up the car at Anaheim Bus Station on March 12th?
\\
    \flan &  \color{orange} What time would you like to pick it up? \\ 
    \opt & \color{orange} What time would you like to pick it up? \\

    Turn & 8 \\ 
    Gold & ApiCall(method=`GetCarsAvailable', parameters=`car\_type': `SUV', `city': `Anaheim', `end\_date': `2019-03-13', `pickup\_time': `12:00',`start\_date': `2019-03-12' )\\ 
    \gptt & \color{ForestGreen} ApiCall(method=`GetCarsAvailable', parameters={`city': `Anaheim', `start\_date': `2019-03-12', `pickup\_time': `12:00', `end\_date': `2019-03-13', `car\_type': `SUV'}) \\ 
    \gptf & \color{BrickRed} Thank you for the information. Could you also specify which type of SUV you would prefer for your rental in Anaheim, or should I proceed with any available SUV?
 \\
    \gpt & \color{ForestGreen}ApiCall(method=`GetCarsAvailable', parameters=\color{BrickRed}`dropoff\_date': `2019-03-13', `pickup\_city': `Anaheim', `pickup\_date': `2019-03-12', \color{ForestGreen}`pickup\_time': `12:00')\\
    \flan & \color{orange}ApiCall(method=`GetCarsAvailable', parameters=`car\_type': `Suv', `city': `Anaheim', `pickup\_date': `2019-03-12', `pickup\_time': `12:00' ) \\ 
    \opt & \color{ForestGreen}ApiCall(method=`GetCarsAvailable', parameters=\color{BrickRed}`car\_type': `Full-size', \color{ForestGreen}`start\_date': `2019-03-12', `end\_date': `2019-03-13', \color{BrickRed}`pickup\_location': `Anaheim', `start\_time': `12:00') \\

    Turn & 9 \\ 
    User & I want to pick it up at 12 in the afternoon and drop it off on the 13th of this month. \\ 
    Gold & There is a nice SUV Dacia Duster available at Anaheim Intermodal Center.  \\ 
    \gptt & \color{BrickRed} When would you like to pick up the SUV rental car in Anaheim on the 12th of this month? \\ 
    \gptf & \color{ForestGreen} We have an SUV available for your dates in Anaheim. It's a Dacia Duster, and you can pick it up at the Anaheim Intermodal Center at 12:00 pm on the 12th of March, and drop it off on the 13th of March. The rental price is \$43.00 per day. Would you like to go ahead and reserve this car? Additionally, do you require insurance for the rental?
 \\
    \gpt & \color{ForestGreen}There is a nice car at Anaheim Intermodal Center for pickup on March 12th. It is a SUV and it is a nice car.
\\
    \flan & \color{ForestGreen}There is a nice car available at Anaheim Intermodal Center on March 12th. It is a Dacia Duster. \\ 
    \opt & \color{BrickRed}There are 5 cars available.\color{ForestGreen} There is a nice Dacia Duster that is available at Anaheim Intermodal Center and it is a SUV. It would be available on March 12th. \\

    Turn & 10 \\ 
    User & How much does it cost daily? \\ 
    Gold & It cost \$43.00 a day.  \\ 
    \gptt & The SUV rental car you requested, a Dacia Duster, is available for pick up at the Anaheim Intermodal Center at 12:00 on March 12th. The rental period will be until 12:00 on March 13th, and the \color{ForestGreen}price per day is \$43.00. I will proceed with reserving this car for you. \\ 
\gptf &  \color{ForestGreen} The Dacia Duster SUV is available for a daily rate of \$43.00. Would you like to proceed with the reservation?
\\
    \gpt & \color{ForestGreen}The cost is \$43.00.
 \\

    \flan & \color{ForestGreen}It would cost \$43.00 per day. \\ 
    \opt & \color{ForestGreen}It costs \$43.00 daily. \\

    Turn & 11 \\ 
    User &  Go ahead and reserve the car \\ 
    Gold &  Would you like to add insurance? \\ 
    \gptt & \color{BrickRed}The SUV Dacia Duster in Anaheim Intermodal Center costs \$43.00 per day. \\ 
  \gptf & \color{ForestGreen} Would you like to add insurance to your car reservation?
 \\
    \gpt & \color{BrickRed}You would like to pick up a nice car at Anaheim Intermodal Center on March 12th at 12 pm and drop it off on March 13th?
\\

    \flan & \color{BrickRed} You would like to pick up a nice car from Anaheim Intermodal Center on March 12th at 12 pm and drop it off on March 13th. Is this correct? \color{ForestGreen} You would like to add insurance to the car. \\ 
    \opt & \color{BrickRed} You want a full-size car and would like to pick it up at Anaheim Intermodal Center on March 12th at 12 pm and drop it off on March 13th. Is that right? \\

    Turn & 12 \\ 
    User & I don't need it. \\ 
    Gold & You would like to pick up a SUV at Anaheim Intermodal Center on March 12th at 12 pm and drop it off on March 13th  \\ 
    \gptt & \color{BrickRed} ApiCall(method=`ReserveCar', parameters={`car\_type': `SUV', `city': `Anaheim', `end\_date': `2019-03-13', `pickup\_location': `Anaheim Intermodal Center', `pickup\_time': `12:00', `start\_date': `2019-03-12', `add\_insurance': `False'}) \\ 
    \gptf & \color{BrickRed} ApiCall(method=`ReserveCar', parameters={`pickup\_location': `Anaheim Intermodal Center', `start\_date': `2019-03-12', `pickup\_time': `12:00', `end\_date': `2019-03-13', `car\_type': `SUV', `add\_insurance': `False'})

 \\
    \gpt & \color{BrickRed}Is there anything else I can help you with?
\\
    \flan & \color{BrickRed} Is there anything else I can help you with? \\ 
    \opt & \color{orange}Would you like to pick up the car at Anaheim Intermodal Center on March 12th at 12 pm? \\

    Turn & 13 \\ 
    Gold & ApiCall(method=`ReserveCar', parameters=`add\_insurance': `False', `car\_type': `SUV', `end\_date': `2019-03-13', `pickup\_location': `Anaheim Intermodal Center', `pickup\_time': `12:00',`start\_date': `2019-03-12' ) \\ 
    \gptt & \color{ForestGreen} ApiCall(method=`ReserveCar', parameters={`car\_type': `SUV', `city': `Anaheim', `end\_date': `2019-03-13', `pickup\_location': `Anaheim Intermodal Center', `pickup\_time': `12:00', `start\_date': `2019-03-12', `add\_insurance': `False'}) \\ 
    \gptf & \color{ForestGreen} ApiCall(method=`ReserveCar', parameters={`pickup\_location': `Anaheim Intermodal Center', `start\_date': `2019-03-12', `pickup\_time': `12:00', `end\_date': `2019-03-13', `car\_type': `SUV', `add\_insurance': `False'})

 \\
    \gpt & \color{ForestGreen}ApiCall(method=`ReserveCar', parameters=`car\_type': `SUV', \color{BrickRed}`dropoff\_date': `2019-03-13', `pickup\_date': `2019-03-12', \color{ForestGreen}`pickup\_location': `Anaheim Intermodal Center', `pickup\_time': `12:00')\\
    \flan & \color{ForestGreen}ApiCall(method=`ReserveCar', parameters=`car\_type': 'SUV', \color{BrickRed}`dropoff\_date': `2019-03-13', \color{ForestGreen}`pickup\_location': `Anaheim Intermodal Center', \color{BrickRed}`pickup\_date': `2019-03-12', \color{ForestGreen}`pickup\_time': `12:00' ) \\ 
    \opt & \color{ForestGreen}ApiCall(method='ReserveCar', parameters={'car\_type': 'SUV', 'end\_date': '2019-03-13', 'pickup\_location': 'Anaheim Intermodal Center', 'pickup\_time': '12:00', \color{BrickRed}'end\_date': '2019-03-13'}) \\

    Turn & 14 \\ 
    User & Yes, that would be great. \\ 
    Gold & I have reserved the car. \\ 
    \gptt & \color{BrickRed} ApiCall(method=`ReserveCar', parameters={`car\_type': `SUV', `pickup\_location': `Anaheim Intermodal Center', `start\_date': `2019-03-12', `end\_date': `2019-03-13', `pickup\_time': `12:00', `add\_insurance': `False'}) \\ 
    \gptf & \color{ForestGreen} Your reservation has been confirmed for the Dacia Duster SUV at Anaheim Intermodal Center. You can pick it up on March 12th at 12:00 PM and return it by March 13th. The cost is \$43.00 per day, and you've chosen not to add insurance. Is there anything else I can assist you with today?
 \\
    \gpt & \color{ForestGreen}I have reserved the car.
\\
    \flan &\color{ForestGreen} Your car has been reserved. \\ 
    \opt & \color{ForestGreen}I have reserved the car. \\

    Turn & 15 \\ 
    User & Thanks a lot. \\ 
    Gold & Anything else you need? \\ 
    \gptt & \color{BrickRed} System: You would like to pick up a SUV at Anaheim Intermodal Center on March 12th at 12 pm and drop it off on March 13th. Your reservation for a Dacia Duster SUV at \$43.00 per day has been confirmed. \\ 
    \gptf &  \color{ForestGreen} You're welcome! If you need any more assistance or have further questions, feel free to ask. Have a great trip!
\\
    \gpt &\color{ForestGreen} Is there anything else?
\\
    \flan & \color{ForestGreen}Is there anything else I can help you with? \\ 
    \opt &\color{ForestGreen} Is there anything else I can help you with? \\

    Turn & 16 \\ 
    User & No, thanks a lot \\ 
    Gold & Enjoy your day. \\ 
    \gptt & \color{ForestGreen}You're welcome! If you need any more assistance or have any other questions, feel free to ask. Have a great day! \\ 
    \gptf & \color{ForestGreen} You're welcome! Have a great trip to Anaheim and enjoy your ride! If you need anything else in the future, just let me know. Safe travels!
 \\
    \gpt & \color{ForestGreen}Enjoy your day.
\\
    \flan & \color{ForestGreen}Have a great day. \\ 
    \opt &\color{ForestGreen} Enjoy your day. \\

\end{longtblr}

\twocolumn

\end{document}